\makeatletter
\disable@package@load{breakurl}{}
\makeatother

\documentclass[default,iicol]{sn-jnl}

\usepackage{graphicx}%
\usepackage{multirow}%
\usepackage{amsmath,amssymb,amsfonts}%
\usepackage{amsthm}%
\usepackage{mathrsfs}%
\usepackage[title]{appendix}%
\usepackage{xcolor}%
\usepackage{textcomp}%
\usepackage{manyfoot}%
\usepackage{booktabs}%
\usepackage{algorithm}%
\usepackage{algorithmicx}%
\usepackage{algpseudocode}%
\usepackage{listings}%

\usepackage{amsfonts}
\usepackage{booktabs}
\usepackage{mathrsfs}
\usepackage{amsthm}
\usepackage{multirow}
\usepackage{cuted}
\usepackage{animate}
\usepackage{subfiles}
\usepackage{color}
\usepackage{amssymb}
\usepackage{ragged2e}
\usepackage{amsmath}
\usepackage{makecell}
\usepackage{array}
\usepackage{multirow}
\usepackage{booktabs}
\usepackage{enumitem}
\usepackage{graphicx}
\usepackage{float}
\usepackage{subfig}

\newcommand{\etal}{\textit{et al}.}
\newcommand{\ie}{\textit{i}.\textit{e}.}
\newcommand{\eg}{\textit{e}.\textit{g}.}

\usepackage{amsthm}

\newcommand{\revise}[1]{\textcolor{black}{#1}}

\theoremstyle{req}

\theoremstyle{thmstyleone}%
\theoremstyle{thmstyletwo}%

\theoremstyle{thmstylethree}%

\begin{document}

\title[Self-Supervised Skeleton-Based Action Representation Learning: A Benchmark and Beyond]{Self-Supervised Skeleton-Based Action Representation Learning: A Benchmark and Beyond}


\author[]{\fnm{Jiahang} \sur{Zhang}}\email{zjh2020@pku.edu.cn}

\author[]{\fnm{Lilang} \sur{Lin}}\email{linlilang@pku.edu.cn}

\author[]{\fnm{Shuai} \sur{Yang}}\email{williamyang@pku.edu.cn}

\author[]{\fnm{Jiaying} \sur{Liu}}\email{liujiaying@pku.edu.cn}

\affil[]{\orgdiv{Wangxuan Institute of Computer Technology}, \orgname{Peking University}, \orgaddress{\city{Beijing}, \postcode{100871}, \country{China}}}


\abstract{Self-supervised learning (SSL), which aims to learn meaningful prior representations from unlabeled data, has been proven effective for skeleton-based action understanding.
Different from the image domain, skeleton data possesses sparser spatial structures and diverse representation forms, with the absence of background clues and the additional temporal dimension, presenting new challenges for spatial-temporal motion pretext task design.
Recently, many endeavors have been made for skeleton-based SSL, achieving remarkable progress.
However, a systematic and thorough review is still lacking. 
In this paper, we conduct, for the first time, a comprehensive survey on \textit{self-supervised skeleton-based action representation learning}.
%
Following the taxonomy of context-based, generative learning, and contrastive learning approaches, we make a thorough review and benchmark of existing works and shed light on the future possible directions.
Remarkably, our investigation demonstrates that most SSL works rely on the single paradigm, learning representations of a single level, and are evaluated on the action recognition task solely, which leaves the generalization power of skeleton SSL models under-explored.
To this end, a novel and effective SSL method for skeleton is further proposed, which integrates versatile representation learning objectives of different granularity, substantially boosting the generalization capacity for multiple skeleton downstream tasks.
Extensive experiments under three large-scale datasets demonstrate our method achieves superior generalization performance on various downstream tasks, including \textit{recognition, retrieval, detection, and few-shot learning}.}




\maketitle

\section{Introduction}
\label{sec:intro}

Human activity understanding is an essential topic in the research of computer vision due to its wide applications in real life, such as human-robotics interaction~\cite{lee2020real}, autonomous driving~\cite{camara2020pedestrian}, and healthcare~\cite{lopez2019human}.
Among the different data modalities for actions, skeletons represent the human body by 3D coordinates of key body joints, which are lightweight, compact, and more robust to changes of view and background. Owing to these desirable advantages, skeleton has attracted much attention in human action analysis.

In the early works, many endeavors have been put into the supervised skeleton-based human activity understanding, \eg, recognition and detection~\cite{yan2018spatial,shi2019two,cheng2020skeleton,rao2024hierarchical}. However, these supervised methods heavily rely on huge amounts of labeled data, which requires time-consuming and expensive data annotation work, limiting the wide applications in the real world. 
As a remedy to this problem,
self-supervised learning (SSL) attracts much attention and has been proven successful for representation learning. It exploits supervisory signals from unlabeled data, learning meaningful prior features and boosting generalization capacity of model for downstream tasks. Motivated by recent success in the image domain, great
interest has arisen in adopting SSL for skeleton. However, it is not trivial to transfer these approaches into the skeleton data directly, which are with a more compact spatial structure, additional temporal dimension, and the absence of the background clues. To this end, researchers have made valuable exploration for skeleton-based SSL.

Generally, existing skeleton-based SSL works can be categorized into three types according to the pre-training pretext tasks, \ie, \textit{context-based}, \textit{generative learning}, and \textit{contrastive learning} methods. The context-based methods construct the pseudo-label based on the intrinsic property of data, \eg, the joint angle prediction, to learn the spatial and temporal relations. Generative learning mainly focuses on reconstructing and predicting the skeleton data or the corresponding features. Contrastive learning methods model the high-level representations with an instance discrimination task. The various positive and negative skeleton views are generated by the well-designed spatial-temporal augmentations, boosting the consistency learning of the model. 

Despite the huge progress recently, there is still a lack of a thorough literature review and analysis.
Therefore, in this paper, we contribute a comprehensive survey of the self-supervised skeleton-based action representation learning. In contrast to other SSL surveys towards image, video, or text data, we focus on skeleton-based representation learning, which is the first literature to the best of our knowledge. This survey conducts a thorough review of mainstream SSL literature for skeleton and also involves the skeleton data collection, benchmark of performance, and the discussion of future possible directions. We believe our extensive work can benefit the research community and bring rich insights for future work.



Based on our literature review, it is noticed that most SSL methods for skeleton focus on the single paradigm, learning representations of single granularity,~\eg, joint-level features (by masked skeleton modeling, MSM)~\cite{zheng2018unsupervised,wu2022skeletonmae} or sequence-level features (by contrastive learning)~\cite{li20213d,mao2022cmd,hico2023,zhang2022hiclr}. This limits the generalization capacity of the model to more downstream tasks. 
Although some works~\cite{lin2020ms2l,su2020predict,wang2022contrast} make valuable efforts to combine different paradigms, they only achieve mediocre improvement due to the inherent gap of feature modeling mechanisms between the contrastive learning and masked modeling~\cite{qi2023contrast,gui2023survey}.
To this end, a novel SSL approach for skeleton is proposed to fully boost the generalization capacity of SSL model, which integrates the contrastive learning and MSM to learn the joint, clip, and sequence level representations jointly. Specifically, we fully utilize the novel motion pattern exposed by the manual designed augmentations and model training for sequence-level contrastive representation learning, while adopting MSM for the joint-level feature modeling. 
Besides, we further propose a novel clip-level contrastive learning method, which significantly boosts the short-term modeling capacity, along with an effective post-distillation strategy to achieve a more compact representation space.
Finally, extensive experiments under \textit{five} downstream tasks, not limited to the single action recognition task used in most previous works, demonstrate the promising generalization capacity of the proposed method.


Our contributions can be summarized as follows:

\begin{itemize}[leftmargin=2em]

\item To the best of our knowledge, we are the first to provide a thorough survey that comprehensively reviews the self-supervised skeleton action representation learning literature. Based on the taxonomy of context-based, generative learning, and contrastive learning, we give a detailed analysis of the pretext task design and highlight the special consideration for skeleton data along with the corresponding challenges. 

\item Motivated by the limitations revealed by our survey, we explore skeleton-based versatile action representation learning to fully mine the generalization power of SSL models. An effective SSL schema is proposed, which integrates contrastive learning and MSM to jointly model the representations of different granularity, remarkably benefiting different downstream tasks. 

\item We present the \revise{first multi-task benchmark} of existing skeleton SSL works with the insightful analysis from the perspective of model backbones, pre-training paradigms, and future possible directions.
Finally, we demonstrate promising performance of our proposed method on \textit{five} downstream tasks for versatile action representation learning.

\end{itemize}


The remaining sections are organized as follows:
{We first present a thorough review in Sec.~\ref{sec:survey}, for skeleton-based SSL representation learning.
Subsequently, based on our investigation, we propose a new method exploring the combination of contrastive learning and masked modeling tasks in Sec.~\ref{sec:method}. Then, we comprehensively benchmark existing methods in Sec.~\ref{sec:survey:evaluation} \revise{on multiple downstream tasks}, and verify the effectiveness of our proposed approach. Finally, we conclude and summarize with possible future directions in Sec.~\ref{sec:conclusion}.}

\section{Review on Skeleton-Based Action Representation SSL}
\label{sec:survey}
\begin{figure*}[t]
    \centering
    \includegraphics[width=0.95\textwidth]{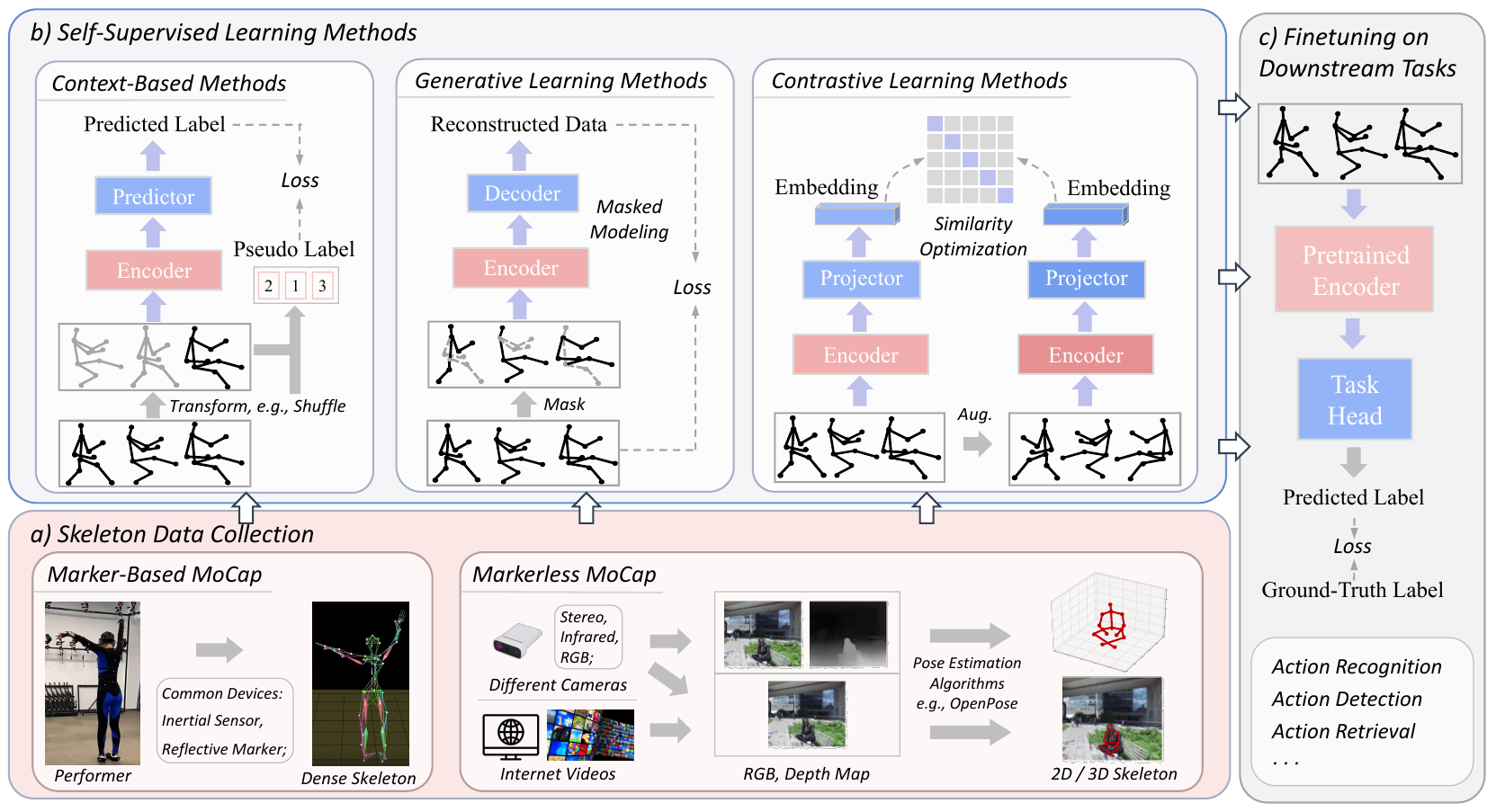}
   \caption{
   The taxonomy framework for self-supervised skeleton-based representation learning in our survey. The survey is structured around three
dimensions: skeleton data collection, SSL pretext design, and SSL downstream task evaluation, providing a comprehensive review.  
   }
   \vspace{-5pt}
  \label{fig:overall}
\end{figure*}

Generally, a two-stage paradigm is utilized in skeleton-based SSL, \ie, pre-training on pretext tasks first and then fine-tuning on downstream tasks. In the pre-training stage, different pretext tasks are designed for deep neural networks, capturing training signals derived from the data itself, called the process of \textit{self-supervision}~\cite{shurrab2022self}. After that, the learned knowledge as feature representations is transferred to downstream tasks as shown in Fig.~\ref{fig:overall} (c). In principle, this part is not only the goal of SSL, \ie, to improve downstream task performance with learned representations, but also the way to assess the quality of representation learning methods. Note that some downstream tasks, \eg, motion prediction and 2D-to-3D lifting, are not considered in this survey, because they essentially do not rely on the supervisory signal of human annotation and can serve as pre-training pretext tasks themselves, leading to possible unfair comparison.
For reviews on these topics, we direct readers to~\cite{lyu20223d,liu2022recent}.

Next, we first introduce skeleton data as well as its collection in Sec.~\ref{sec:survey:data_collect}. Then, a review of the skeleton SSL methods is presented in Sec.~\ref{sec:survey:method_review} based on the taxonomy of pretext tasks methodologies as shown in Fig.~\ref{fig:overall} (b). A summary and discussion are finally provided in Sec.~\ref{sec:survey:discuss}. 

\subsection{Human Skeleton Representation}
\label{sec:survey:data_collect}
The collection methods of skeleton can be divided into two main categories, \ie, marker-based and markerless methods, shown in Fig.~\ref{fig:overall} (a).

\noindent\textbf{Marker-based} motion capture (Mocap) systems often rely on inertial measurement units (IMU) or reflective markers, 
placed on the body to track the movement of humans. It can provide reliable skeleton data with even sub-millimeter accuracy~\cite{buckley2019role}. However, this method is too costly in many application scenarios and requires highly trained personnel to operate. Meanwhile, it suffers from great inconvenience, \eg, the time-consuming placement process and a requirement for a controlled environment. The datasets~\cite{ionescu2013human3,von2018recovering} for generation tasks, \eg, pose estimation, are usually collected in this way to guarantee the accuracy. 

\noindent\textbf{Markerless} Mocap often depends on deep learning algorithms for pose estimation from RGB and depth data. 
Video RGB data can be easily obtained from the Internet while we can also utilize the camera hardware, \eg, the depth camera including Microsoft Azure Kinect, 
and single or multiple RGB video cameras, to collect the RGB, infrared, and depth images in deployment. Then to get the 3D motion, computer vision algorithms, on multi-view
geometry and pose estimation, \ie, \textit{OpenPose}~\cite{cao2017realtime} are then employed to detect and extract joint center locations. The whole pipeline is presented in Fig.~\ref{fig:overall} (a). However, due to the limitations of hardware and estimation algorithms, such method can have large errors compared to marker-based methods~\cite{dolatabadi2016concurrent}. Nevertheless, it is still chosen for most skeleton-based  datasets~\cite{shahroudy2016ntu,liu2019ntu,liu2020pku} for its simpleness and convenience.

\subsection{SSL Methods for Skeleton}
\label{sec:survey:method_review}
\revise{SSL has been widely explored in the previous literature for the image domain. Some early works adopt the context-driven pretext tasks to model the inherent data structure, \eg, predicting the rotation label~\cite{gidaris2018unsupervised}, correlation map~\cite{li2023correlational}, and clusters based on the pseudo-labels~\cite{caron2018deep,zhan2020online}. Besides, contrastive-style methods push the performance of SSL to new heights, including invariance decorrelation~\cite{zbontar2021barlow,bardes2021vicreg},
distillation-based~\cite{grill2020bootstrap, chen2021exploring,caron2021emerging,oquab2023dinov2,darcet2023vision}, clustering-based~\cite{caron2020unsupervised,li2020prototypical}. Meanwhile, the generative learning, especially the recent masked image modeling~\cite{xie2022simmim,he2022masked}, has drawn a surge of interest and been proven effective for downstream vision downstream tasks. These works have played an important role and motivate the new efforts for skeleton-based SSL. However, due to the huge difference between human skeleton and image data as discussed in Sec.~\ref{sec:intro}, it is suboptimal to transfer these methods directly for skeleton representation learning. Therefore, we present this literature review with a high-level methodological taxonomy, and also shed light on the challenges and skeleton-specific designs to motivate further work.}

\begin{figure*}[t]
    \centering
    \includegraphics[width=0.92\textwidth]{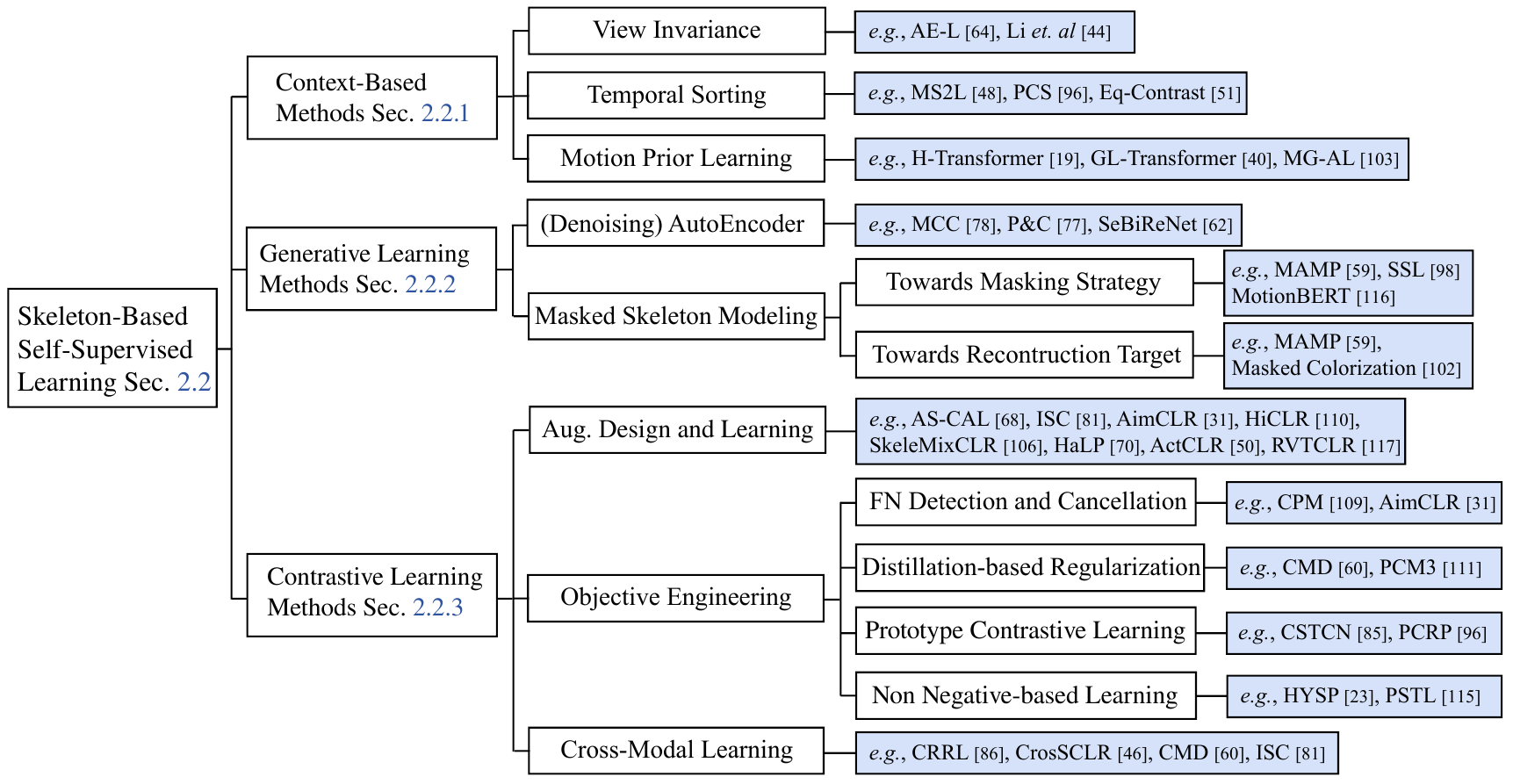}
   \caption{
   The taxonomy of the skeleton-based self-supervised learning methods in our review. 
   }
   \vspace{-5pt}
  \label{fig:structure}
\end{figure*}

With respect to the pretext tasks, most existing skeleton-based representation learning methods encompass three categories: (1) \textit{context-based}, (2) \textit{generative learning}, and (3) \textit{contrastive-learning} methods. Based on the taxonomy in Fig.~\ref{fig:structure}, we provide a comprehensive survey on the skeleton-based SSL works and highlight the special design for the skeleton as well as the corresponding challenges, to distinguish them from other data modalities.

\subsubsection{Context-Based Methods}
\begin{figure}
    \centering
    \subfloat[View invariance learning.]{
       \includegraphics[width=0.99\linewidth]{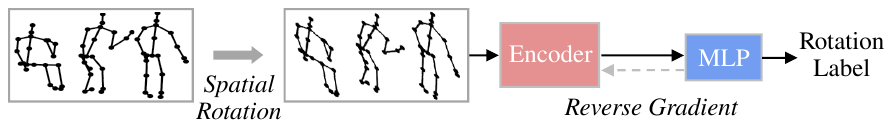}}

  \vspace{-7pt}
  \subfloat[Temporal sorting.]{
       \includegraphics[width=0.99\linewidth]{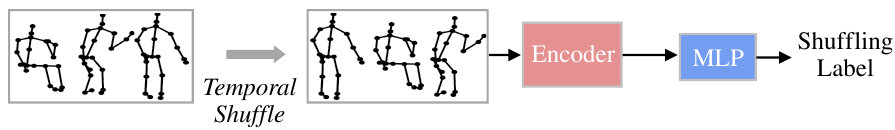}}
  
  \vspace{-7pt}
  \subfloat[Motion prior knowledge learning.]{
       \includegraphics[width=0.99\linewidth]{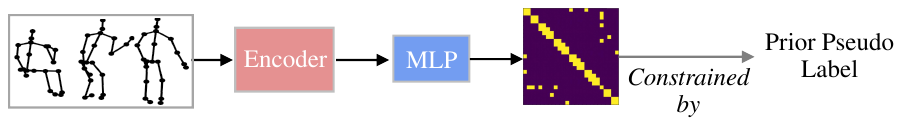}}
    \caption{Three types of context-based SSL methods for skeleton.}
    \vspace{-5pt}
    \label{fig:context}
\end{figure}

Context-based methods generate the supervisory training signals according to the inherent contextual information of provided data. The model is encouraged to learn the spatial-temporal relationships by training on the pre-defined task. Emerging from the image domain, the pretext tasks rely on the context understanding, \eg, rotation prediction~\cite{gidaris2018unsupervised}. In contrast, skeleton data introduces an additional temporal dimension, and possesses more compact spatial information, which presents a new challenge on \textit{how to mine the meaningful spatial-temporal context of skeleton by pretext tasks}.
Typically, there are three common types of context-based pretext tasks, {view-invariance-based}, {temporal-order-based}, and {motion-prior-based} method as shown in Fig.~\ref{fig:context}.

\vspace{0.3em}
\noindent\textbf{1) View Invariance.} Due to the variance of the observation viewpoint, the skeleton estimation can suffer from occlusion and noise. Therefore, learning view-invariant representations should be beneficial for action recognition task, and has been widely studied in the supervised skeleton-based action recognition task~\cite{gao2022global}. Li~\etal~\cite{li2018unsupervised} proposed a view classification pretext task for unsupervised action representation learning. Specifically, as shown in Fig.~\ref{fig:context} (a), skeleton sequences of different views are fed into the encoder, subsequent to which a view classifier predicts the view labels. To learn the view-invariant features, a Gradient Reversal Layer~\cite{ganin2015unsupervised} is added to reverse the optimization direction of the encoder. Resort to this, the encoder can learn the features insensitive to view in an adversarial manner. Likewise, Paoletti~\etal~\cite{paoletti2022unsupervised} adopted the rotation prediction, \ie, pitch angle, yaw angle, and roll angle, as well as a gradient reverse operation to achieve the viewpoint-invariance learning.

There are also some subsequent SSL works~\cite{yang2022via,men2023focalized,bian2022view} focusing on learning view-invariant representations. However, they adopted a contrastive learning paradigm to explicitly learn the alignment between the two views and showed a significant performance improvement.

\vspace{0.3em}
\noindent\textbf{2) Temporal Sorting.}
In this pretext task, skeleton data is treated as temporal sequences, and shuffled randomly. The model takes the shuffled skeleton as input and outputs the corresponding shuffling pseudo label to restore the temporal order. 
Specifically, the skeleton sequence is usually divided into multiple clips first, each of which contains several consecutive frames. Then the shuffling operation takes place at the clip level rather than the frame level because it is difficult to capture motion patterns by the difference between two adjacent frames.
The model is trained to predict the shuffling label as shown in Fig.~\ref{fig:context} (b), \eg, with a \textit{Cross-Entropy} loss.

Some skeleton-based SSL works~\cite{lin2020ms2l,xu2021prototypical} are equipped with temporal sorting to model the temporal dependencies, similar to SSL works in video~\cite{xu2019self,lee2017unsupervised}. However, such pretext task is often integrated with other pretext tasks, because it only models temporal features explicitly, leaving the crucial spatial representations of skeleton under-explored. Therefore, new pretext tasks are studied to jointly learn the spatial-temporal relationships of skeleton.

\vspace{0.3em}
\noindent\textbf{3) Motion Prior.}
The motion dynamics of skeleton joints contain rich semantic information beneficial for action understanding. Therefore, researchers propose to leverage the inherent motion prior knowledge to generate pseudo labels for pre-training as shown in Fig.~\ref{fig:context} (c). Cheng~\etal~\cite{cheng2021hierarchical} designed a movement direction prediction task. The model is constrained to estimate the direction of the instantaneous joint velocity, \ie, whether the targeted joints are moving in a positive direction. Based on this, Kim~\etal~\cite{kim2022global} proposed a multi-interval pose displacement strategy to encourage both local and global attention learning. The key idea is to predict the motion direction and magnitude of the central joint and other joints during a random interval. 
On the other hand, Yang~\etal~\cite{yang2022motion} introduced more motion prior information as the pseudo label, \ie, intra-joint motion variance, inter-joint motion covariance, intra-frame joint angle, and inter-frame motion deviation. The model is pre-trained by a regression task to predict this prior knowledge.

\vspace{0.5em}
In summary, compared with image and video modalities, the context-based pretext tasks for skeleton usually involve more spatial-temporal design and motion dynamics modeling. However, these tasks are often designed manually and hard to capture the high-level underlying semantic distribution. Therefore, more efforts have been paid to the generative and contrastive learning SSL methods, which will be discussed in the following.

\subsubsection{Generative Learning Methods}
Generative methods utilize the generative power of the neural network to capture the spatial-temporal co-occurrence relationships among skeleton joints, modeling the underlying data distributions.
In this case, the meaningful representations are learned by reconstructing or predicting the related signals of input skeletons. Here we do not strictly distinguish between ``reconstruction" and ``prediction" for generation, but use ``reconstruct" to refer uniformly. Generally, the goal of the generative learning tasks can be formulated as:
\begin{equation}
    \centering
    \label{eq:recons}
    argmin_{\theta} \ \mathcal{L}(\mathcal{D}(\mathcal{E}(\mathcal{T}_{in}(x))), \mathcal{T}_{tar}(x); \theta),
\end{equation}
where $x$ is a skeleton sequence, $\mathcal{D}$ and $\mathcal{E}$ are the decoder and encoder, respectively. $\mathcal{T}_{in}$ is the transformation applied to the original data, \eg, random masking, while $\mathcal{T}_{tar}$ maps the input into the target space where the loss objective $\mathcal{L}$ is applied to optimize the model parameters $\theta$. 

When $\mathcal{T}_{in}$ and $\mathcal{T}_{tar}$ are both the identity function, the model is constrained to reconstruct the input data simply, known as AutoEncoder (AE). It naturally creates an information bottleneck to achieve dimension reduction, mapping data from input space onto low-dimensional feature space. Due to its simpleness, AE is used to learn prior representations in earlier skeleton-based works~\cite{su2020predict,xu2021unsupervised}. However, since the skeleton data is still redundant in spatial and temporal dimensions, the encoder tends to memorize the input data at a low level instead of modeling the high-level semantic knowledge. Then, denoising AE (DAE) is studied for alleviating this shortcut~\cite{su2021self, nie2020unsupervised}. For example, Nie \etal~\cite{nie2020unsupervised} applied a series of view corruptions and constrained the model to reconstruct the clean data to disentangle the view and pose features. On the other hand, inspired by the remarkable success of masked language/image modeling (MLM, MIM), which can be regarded as a variant of DAE, researchers have aroused a surge of interest in exploring masked skeleton modeling (MSM) for human action representation learning. The earlier works are mainly based on recurrent neural network (RNN) or graph convolutional network (GCN) to predict the masked skeletons. However, lacking good scalability, these models only show mediocre performance improvement. Recently, inspired by the success of Vision Transformer (ViT) as masked autoencoders (MAE)~\cite{he2022masked}, Transformer-based models have been explored for the masked skeleton modeling~\cite{chen2022hierarchically,mao2023masked}. However, due to the relative redundancy in skeleton and the lack of a large-scale dataset as ImageNet, directly applying Transformer for masked skeleton reconstruction can suffer from the over-fitting problem. To this end, researchers have made endeavors on two crucial designs, masking strategy and reconstruction target, which are discussed as follows.

\vspace{0.3em}
\noindent\textbf{1) Masking Strategy} has been proved crucial as the design of $\mathcal{T}_{in}$ in Eq.~(\ref{eq:recons}) for MIM. In the field of RGB images, MAE~\cite{he2022masked} adopts a patch-based masking strategy with a large masking ratio of 75\%. For the skeleton data, researchers have fully explored the masking strategy in the spatial-temporal dimension. Although the optimal masking strategy and ratio can differ with the encoder backbone and MSM task setting, most works have found that spatially body-part-based and temporally segment-based strategy with large masking ratio, \eg, 90\% in~\cite{mao2022cmd}, can produce decent results. Concretely, body-part level masking refers to performing masking regarding the different joints in a body part as a whole, \eg, \textit{torso} and \textit{left leg}, while the temporal segment-based strategy masks the same joints across consecutive frames. These designs aim to reduce the shortcuts caused by spatial-temporal redundancy in skeleton, which is in line with the observation in images~\cite{he2022masked} and videos~\cite{tong2022videomae}.

Recently, more elaborate masking strategies have been studied to achieve more effective exploitation of valuable semantic information. The work~\cite{yan2023skeletonmae} found that masking the limbs is always better than masking the torso and head, especially the right hand and leg. MAMP~\cite{mao2023masked} proposes a motion-aware masking strategy. The moving parts are located and masked by calculating the motion displacement between adjacent frames, which achieves better results than the random masking strategy. These results demonstrate that masking motion regions, which are often semantic-rich, promotes the model to learn more meaningful features in masked motion modeling. It can be also explained by the theory in~\cite{zhang2022mask}, \ie, because samples from the same action often contain similar motion patterns, this masking strategy on the motion regions implicitly leads to better alignment of mask-induced positive pairs, achieving more discriminative feature spaces.

Meanwhile, some literature adopts special mask designs, yielding new MSM task forms. MS$\rm ^2$L~\cite{lin2020ms2l} directly utilizes temporal frame-level masks to train the model on the motion prediction task.
MotionBERT~\cite{zhu2023motionbert} adopts a 2D-to-3D lifting pretext task, in which it masks all the values in depth $z$ channel and partially in $x$ and $y$ channels, encouraging the model to predict the original 3D skeletons. These works can be divided into masked skeleton modeling in a broad sense.

\vspace{0.3em}
\noindent\textbf{2) Reconstruction Target}, known as the $\mathcal{T}_{tar}$, also varies from the input space~\cite{he2022masked,xie2022simmim} to the feature space~\cite{assran2023self,gao2024mimic,xie2022masked,haghighat2023pre,bardes2024revisiting} as the image domain.
Most existing works reconstruct the skeleton data in the input coordinate space using MSE loss. 
Some works~\cite{wang2022contrast} propose to perform temporally reverse reconstruction to learn the skeleton dynamics instead of trivial representations that just remember the input. 

Instead of directly reconstructing the input, Mao~\etal~\cite{mao2023masked} proposed to reconstruct the motion in MSM, \ie, the difference of the corresponding joints between adjacent frames. The work~\cite{yang2023self} formulates the skeleton as an unordered 3D point cloud and maps the 3D data onto color space. This mapping function is artificially defined according to the spatial-temporal relationship of skeleton joints. Therefore, the model can learn the spatial relation and temporal dependency by reconstructing the color of the skeleton cloud.

\subsubsection{Contrastive-Learning Methods}
\label{sec:contrastive_learning}
Contrastive learning has been proven effective for different data modalities, \eg, images, point cloud, as well as skeleton data. Generally, contrastive learning pursues the consistency of the \textit{positive samples}, which are usually the augmented counterparts of the original data. MoCo~\cite{he2020momentum} utilizes the negative samples to establish an instance discrimination pretext task. Meanwhile, self-distillation~\cite{grill2020bootstrap} and feature decoupling~\cite{zbontar2021barlow,bardes2021vicreg} concepts have also been explored in contrastive learning. 
These early pioneering works~\cite{he2020momentum,chen2020simple,grill2020bootstrap,zbontar2021barlow,robinson2020contrastive}, 
have made a huge impact and encouraged unique designs for skeleton contrastive learning. 

For skeleton contrastive representation learning, most existing methods are based on MoCo v2~\cite{he2020momentum}. It adopts an asynchronously momentum-updated key encoder and an online query encoder, along with a memory queue to store a large number of consistent negative samples. Specifically, different augmentations are applied to the skeleton $x$ to generate the positive pair $(x_q, x_k)$, while the negative sample features $m_i$ are stored in a memory queue. The model is constrained to retrieve the positives among the negative samples, optimizing the following InfoNCE objective $\mathcal{L}_{Info} (z_q, z_k)$:
\begin{equation}\label{eq:infonce}
\mathcal{L} = -\log\frac{\exp(z_{q} \cdot z_k / \tau)}{\exp(z_{q} \cdot z_k / \tau) + \sum_{i=1}{\exp(z_{q} \cdot m_{i} / \tau)}},
\end{equation}
where the $z_q$/$z_k$ is the query/key embedding encoded by the query/key model and $\tau$ is the temperature hyper-parameter. {$m_{i}$ is the $i$-th feature anchor as the negative sample.} Based on this, we review current skeleton contrastive learning methods from three aspects of design, \ie, \textit{augmentation design and learning strategy, objective engineering}, and \textit{cross-modal learning}. 

\noindent\textbf{1) Augmentation Design and Learning Strategy.}
Data augmentation exposes novel motion patterns and generates diverse positive views, which have been found crucial to the success of contrastive learning~\cite{tian2020makes,guo2022aimclr}. Unlike images, data augmentation for skeleton is relatively less developed. Therefore, the earlier works mainly focus on exploring practical spatial-temporal augmentations for skeleton contrastive learning~\cite{rao2021augmented, thoker2021skeleton, gao2021efficient}. For example, Rao~\etal~\cite{rao2021augmented} proposed a series of augmentations including \textit{Rotation}, \textit{Shear}, \textit{Reverse}, \textit{Gaussian Noise}, \textit{Gaussian Blur}, \textit{Joint Mask} and \textit{Channel Mask}, some of which are used as the default basic augmentations in the future research. Notably, the optimal augmentations for different backbones are usually different. For example, the \textit{Joint Mask} is found detrimental for GCNs~\cite{zhang2022hiclr}, while beneficial for GRU model~\cite{thoker2021skeleton}, which implies that GCNs are more sensitive to the spatial corruption. Generally, this difference arises on account of different modeling mechanisms and model capacities, which further increases the difficulty of the augmentation design and selection.  

Inspired by the success of mixing-based augmentations for images, Chen~\etal~\cite{chen2022skelemixclr} proposed SkeleMixCLR equipped with \textit{SkeleMix} augmentation which combines the topological information of different skeleton sequences. To learn from this mixed skeleton, SkeleMixCLR obtains the corresponding part level and the whole-body level features and pursues the local-global invariance. Based on this, SkeAttnCLR~\cite{hua2023part} further designs an attention mechanism to perform local contrastive learning on salient and non-salient features. Due to the generated novel input views and regularization effect on the feature space, mixing augmentation often leads to consistent improvement for representations.

To further improve consistency learning, researchers have made endeavors to introduce more and stronger data augmentations. However, the ensuing problem is the over-distortion~\cite{bai2022directional} of the augmented data, leading to the model performance degradation. In other words, some strong augmentations would seriously corrupt the semantic information, change the data distribution, and result in the difficulty of model consistency learning. To address this, AimCLR~\cite{guo2022aimclr} utilizes two branches to encode the weakly and strongly augmented views, respectively. Then the model optimizes the similarity of the distributions output by two branches as a soft consistency learning target:
\begin{equation}\label{cond_distri}
\begin{aligned}
    \mathcal{L}_{Soft}(z) &= KL (p(z|z_{weak})\Vert p(z|z_{strong})),\\
	{p}\left(z|z_* \right) &= \frac{\exp(z \cdot z_* / \tau)}{\exp(z_k \cdot z_*/ \tau) + \sum_{i=1}^{M}{\exp(m_{i} \cdot z_* / \tau)}},
\end{aligned}
\end{equation}
where $z_{weak}$ and $z_{strong}$ are the corresponding embeddings of weakly and strongly augmented views. Instead of the one-hot target, this objective utilizes the similarity distribution as the soft target to guide consistency learning, which can be viewed as a self-distillation process. Further, Zhang~\etal~\cite{zhang2022hiclr} introduced more strong augmentations, \eg, randomly dropping the skeleton edges and joints, and proposed a hierarchical contrastive learning framework. It performs a decoupled progressive augmentation invariance learning by optimizing the consistency only between the augmented samples with adjacent strength, where the weakly augmented branch serves as the mimic target of the strongly augmented branch. These works show that the weakly augmented views can effectively guide the learning of the corresponding strongly augmented samples, leading to more stable representation learning and improvement.

In addition, Lin~\etal~\cite{lin2023actionlet} explicitly distinguished the static regions and motion regions, namely, \textit{actionlet}, in human skeleton, and introduced a motion-aware augmentation strategy. By mining the actionlet in an unsupervised manner, the semantic-reserving augmentations are employed for actionlet regions, while the noise perturbations for non-actionlet regions, avoiding the over-distortion problem.
On the other hand, instead of augmenting at the input level, HaLP~\cite{shah2023halp} proposes a latent positive hallucinating method by exploring the latent space around the corresponding prototype.

In summary, the development of data augmentations along with the corresponding learning strategy significantly boosts the performance of skeleton contrastive learning, which has always been an important and popular topic for 3D skeleton contrastive learning.

\noindent\textbf{2) Objective Engineering.} In addition to the widely used InfoNCE loss, some SSL works have also made efforts to explore new loss functions for extra regularization or new objectives. Here we introduce them from the perspective of \textit{False Negative} (FN) problem, which widely exists in skeleton contrastive learning based on negative examples, \eg, MoCo v2. Concretely, false negatives refer to the negative samples but from the same semantic category. Traditional contrastive learning relies on the one-hot label and directly pushes them away in the feature space, which forces the model to discard the shared semantic information and leads to slow convergence~\cite{huynh2022boosting}. We point out that, due to the lack of description of objects and backgrounds, there are fewer action categories represented by skeleton data, which leads to a more serious FN problem. To this end, the following aspects are considered to tackle this issue.

\revise{The first straightforward solution is based on the false negative detection and cancellation~\cite{chen2021incremental,huynh2022boosting,li20213d,guo2022aimclr,zhang2022cpm}. These methods first calculate the similarity between the sample and negatives, selecting the top $k$ negatives that are most similar as the potential false negatives, which are then involved as extended positives in contrastive learning.}

\revise{Some other methods~\cite{xie2022delving,mao2022cmd,zhang2023prompted} suggest employing the distillation objective as an adaptive re-weighting regularization for the one-hot instance discrimination pretext task, which can be formulated as follows}:
\begin{equation}\label{eq:distill}
\begin{aligned}
    \mathcal{L}_{KD}(z_q, z_k) &= -{p}\left({z}_{k},\tau_{k}\right)\log {p}\left(z_{q},\tau_{q}\right),\\	{p_{j}}\left({z},\tau\right) &= \frac{\exp({z} \cdot {m}_{j} / \tau)}{\sum_{i=1}{\exp({z} \cdot {m}_{i} / \tau)}}.
\end{aligned}
\end{equation}
They assign the attraction weights to negatives based on the calculated similarity. Specifically, if a sample possesses high similarity with the positive, a larger attraction weight would be assigned to involve it in similarity optimization.

\revise{Besides, prototype-based contrastive learning is also studied for high-level semantic consistency learning~\cite{li2020prototypical,caron2020unsupervised,xu2021prototypical}. It usually performs clustering in feature space to assign instances to different cluster prototypes as a pseudo-semantic label. Then the model learns more high-level semantics by contrasting different prototypes. By virtue of such a way, the model pays more attention to the cluster-level discrimination task rather than instance-level to alleviate the false negative problem.}

In addition to the above designs, as introduced above, some negative-sample-free contrastive frameworks have been developed recently, which avoid the difficulty of explicitly specifying negative examples. Based on BYOL~\cite{grill2020bootstrap}, the works~\cite{francohyperbolic,chen2023self} performs the positive-only consistency learning in the hyperbolic space. Some other works~\cite{zhang2022skeletal,sun2023unified}, inspired by the feature decorrelation concept, propose to learn the decorrelated representations based on Barlow twins~\cite{zbontar2021barlow} and Variance-Invariance-Covariance Regularization (VICReg)~\cite{bardes2021vicreg} as the framework.

\begin{figure}[t]
    \centering
    \includegraphics[width=0.5\textwidth]{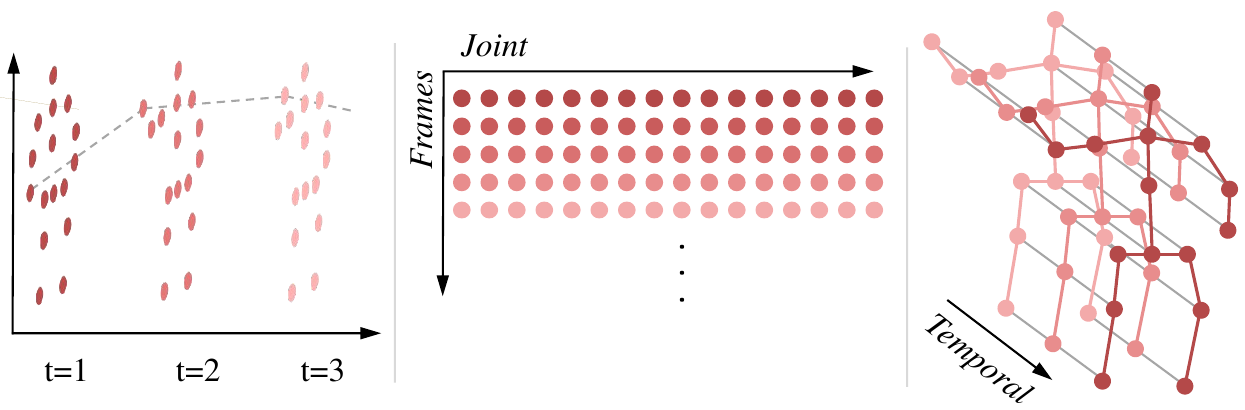}
   \caption{
    Different representations of skeleton data. From left to right are \textit{time series, 2D pseudo-image, spatial-temporal graph}.
   }
   \vspace{-5pt}
  \label{fig:diff_repre}
\end{figure}

\begin{figure*}[t]
    \centering
    \includegraphics[width=0.89\textwidth]{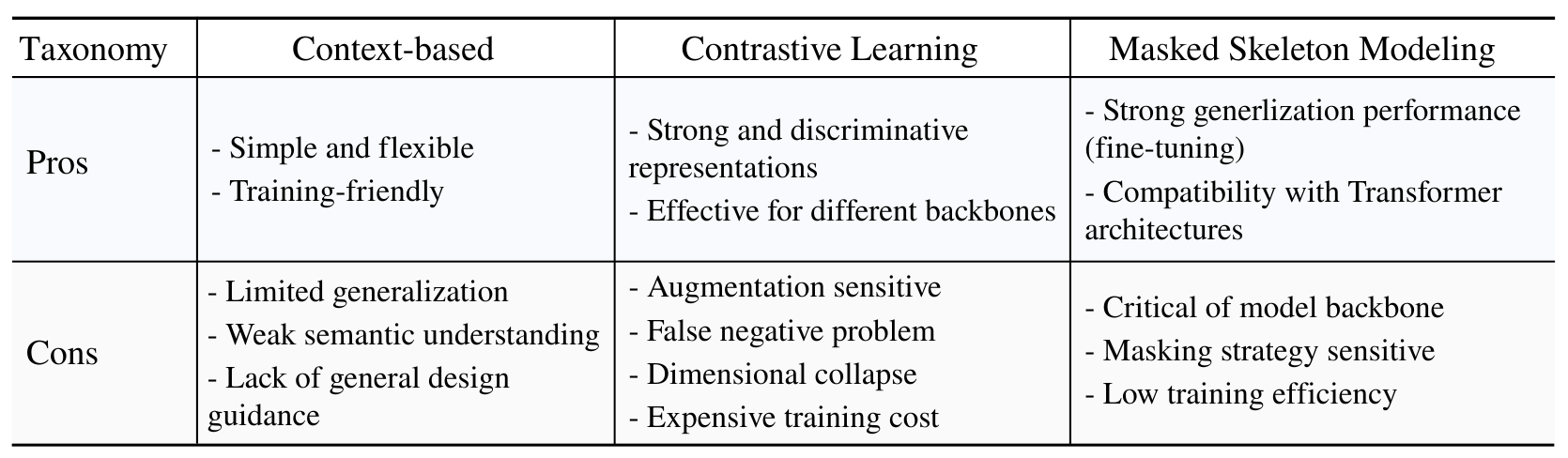}
   \caption{
    \revise{Summary of the advantages/limitations of different skeleton-based SSL methodologies.}}
   \vspace{-5pt}
  \label{fig:ssl_summary}
\end{figure*}

\noindent\textbf{3) Cross-Modal Learning.} Skeleton naturally provides different data modalities to represent human movement, \eg, \textit{motion} and \textit{bone} modality defined in the previous works~\cite{shi2019two,chen2021channel}. Meanwhile, many works~\cite{shi2019two,chen2021channel} have proved that the fusion of different modal knowledge can be beneficial to achieve richer and more representative action depictions. \revise{For example, \textit{joint} contains essential information about human movements while lacking the explicit encoding of some high-order information. \textit{Bone} view can capture the structural patterns well. However, it is subject to the predefined human model and still lacks temporal information. \textit{Motion} explicitly encodes the temporal dynamics while losing the spatial information. It is promising to explore these different modalities to achieve better action representation.} 

CRRL~\cite{wang2022contrast} performs contrastive learning between the joint and motion modalities, where the encoders for motion and joint data are homologous, \ie, obtained by momentum-updated strategy. However, this design can be unreasonable because the model can have difficulty dealing with two modalities simultaneously. 
In contrast, CrosSCLR~\cite{li20213d} adopts three separate encoders for joint, bone, and motion data, and aims to align the distributions among the neighborhoods in latent space. Intuitively, it encourages the sample embedding together with its neighbors of modality $v$ should also be close in the latent space of modality $u$. However, this leads to a two-stage training process, as the model needs to be pre-trained on a single view first to obtain reliable unique modal knowledge. To this end, CMD~\cite{mao2022cmd} proposes a general bidirectional distillation objective, where the modal knowledge is directly modeled as the whole similarity distribution in the customized latent space. The proposed cross-modal distillation objective and the instance discrimination task jointly optimize the model, yielding a concise single training phase. On the other hand, to get rid of the limitation of using separate encoders for different modalities, UmURL~\cite{sun2023unified} develops a unified modality-agnostic encoder, which can handle different modal inputs relying on the modality-specific input embedding layer and feature projection layer.

Meanwhile, we can also organize skeleton data into different representations as shown in Fig.~\ref{fig:diff_repre}, \eg, the temporal series, 2D pseudo-images with frame and joint dimensions, and spatial-temporal graphs with nodes and edges. Following this concept, ISC~\cite{thoker2021skeleton} encodes these different skeleton representations with different model backbones, which are then projected into a shared latent space. A cross-modal contrastive learning loss is applied for pre-training.

\noindent\textbf{4) Others.} In addition to the aforementioned literature, some works also employ contrastive learning. Considering the hierarchical structure of skeleton data, HiCo~\cite{hico2023} performs contrastive learning in a hierarchical manner to model features of different levels. Some works~\cite{lin2020ms2l,xu2021prototypical,tanfous2022and,wang2022contrast} combine the contrastive learning paradigm with others in a multi-tasking manner. Notably, in addition to simply combining reconstruction and contrastive learning pretext tasks, PCM$^{\rm 3}$~\cite{zhang2023prompted} proposes a collaborative design to further improve the representation learning.  Eq-Contrast~\cite{lin2024mutual} formulates the temporal sorting task into an equivariant contrastive learning objective for multi-task pre-training. Meanwhile, not limited to SSL,  contrastive learning is also studied in supervised~\cite{zhou2023learning} and semi-supervised~\cite{xu2022x,shu2022multi} learning for skeleton.

\subsection{Summary and Discussion}
\label{sec:survey:discuss}
The skeleton-based SSL literature is categorized into context-based, generative learning, and contrastive learning types. All these methods aims to capture joint features as well as their relationships from unlabeled skeleton data, to obtain a meaningful representation of the motion. Due to the lack of ground-truth labels, these methodologies introduce prior knowledge in different means to enable representation learning, \eg, context task design, augmentation choices, and masking strategy.
\revise{To summary the different skeleton-based SSL methods based on our taxonomy, we present the advantages and limitations of each methodology category in Fig.~\ref{fig:ssl_summary}. }

\revise{Context-based methods are often simple to implement, flexible for design, and training-friendly compared with the other two paradigms. However, as evidenced by the previous literature, the main concern is the limited model capacity of generalization and semantic understanding due to the gap between the low-level pretext tasks and the downstream tasks. Contrastive learning is recognized by its highly discriminative representations with strong performance, especially under linear evaluation protocol. More importantly, it has been proven effective for different backbones for skeleton modeling. The other side of the coin is augmentation sensitivity, false negative problem, and high training cost as previously discussed. Meanwhile, some researchers point out the concern of dimensional collapse in contrastive learning~\cite{jingunderstanding,lin2024mutual}, where the embedding vectors only span a lower dimensional space, limiting the representation capacity. Finally, the masked skeleton modeling is discussed considering its dominant position in generative learning. It is well compatible with Transformer, with expected strong generalization performance, especially under fine-tuning protocol. However, performance can significantly drop when employing other backbones, which can limit its deployment in some low-latency, resource-constrained scenarios. Meanwhile, it is also concerned about masking strategy sensitivity and time-consuming training.}

\revise{To end this section and possibly motivate future work, we further summarize and emphasize the special design of skeleton-based SSL approaches that go beyond the high-level SSL methodology taxonomy.}

\noindent \revise{$\bullet$ \textit{Context Pretext Tasks Based on Human Topology and Kinematics.} Different from images, skeleton pretext tasks focus on the spatial-temporal data structure learning of human body, \eg, temporal sorting, angle, and displacement prediction.}

\noindent \revise{$\bullet$ \textit{Data Augmentation for Novel Motion Patterns.} Skeleton augmentation has always been the focus to boost contrastive learning. These efforts include the spatial and temporal, invariant and equivariant~\cite{lin2024mutual}, weak and strong~\cite{guo2022aimclr,zhang2022hiclr}, intra and inter augmentations~\cite{chen2022skelemixclr,zhang2023prompted} as well as the corresponding learning strategy.}

\noindent \revise{$\bullet$ \textit{Cross-Modal Knowledge Modeling.} Skeleton naturally
provides different data modalities (or views) to represent
human movement, \eg, joint, motion, and bone, enabling cross-modal knowledge modeling for better action representations.}

\noindent \revise{$\bullet$ \textit{False-Negative Cancellation.} As discussed, the diversity of human motions is much less than natural images, leading to a more serious false negative problem, which is an important topic in skeleton contrastive learning.}

\noindent \revise{$\bullet$ \textit{Masking Strategy in MSM.} As a crucial component in MSM, spatial-temporal masking strategy has also been widely studied specifically for the highly structured human skeleton data.}

\noindent \revise{$\bullet$ \textit{Multi-Granularity Representation Learning.} Skeleton data naturally presents a hierarchy, \eg, sequence-level, clip-level, and joint-level. It can be beneficial to capture these action clues at different scales, enriching the action representation and alleviating the over-fitting problem.}


\section{The Proposed Method}
\label{sec:method}
\subsection{Motivation}
\revise{By reviewing the previous works, we find that most existing skeleton SSL methods employ a single paradigm or simply combine different methods with action recognition task for evaluation solely, which limits the model capacity of learning representations at different granularity and levels and leaves the generalization capacity leashed.
To this end, we propose prompted contrast with masked motion modeling (PCM$\rm ^3$++), which combines contrastive learning and MSM by exploring potential collaboration. It enables versatile representation learning by joint, clip, and sequence level feature modeling, and significantly improves the performance of different downstream tasks.} 

\begin{table}[t]
  \centering
 \setlength{\tabcolsep}{1.0mm}{
  \caption{\revise{Main notations and the definition.}}
  \label{tab:notation}
  \vspace{-5pt}
  \begin{tabular}{l|c}
   \toprule
   Notation & Definition\\
   \midrule
   \multirow{2}{*}{$s^\prime(s_{ach})$, $z^\prime(z_{ach})$} & The global key-branch intra-augmented \\
   & skeleton and its embedding.\\
   \hline
   \multirow{2}{*}{$s_{intra}$, $z_{intra}$} & The query-branch intra-augmented \\
   & skeleton and its embedding.\\
   \hline
   \multirow{2}{*}{$s_{inter}$, $z_{inter}$} & The query-branch inter-augmented \\
   & skeleton and its embedding.\\
   \hline
   \multirow{2}{*}{$s_{clip}$, $z_{clip}$} & The query-branch temporal-augmented \\
   & skeleton clip and its embedding. \\
   \hline
   \multirow{2}{*}{$s_{mask}$, $z_{mask}$} & The query-branch masked \\
   & skeleton and its embedding. \\
   \hline
   \multirow{2}{*}{$s_{predict}$, $z_{predict}$} & The reconstructed skeleton from $s_{mask}$\\
   & and its embedding. \\
   \hline
   \multirow{2}{*}{$z_{inter}^{\prime}$} & The linear interpolation of two skeleton \\
   & embeddings in mix inter-augmentation.\\
   \hline
   $p_*$ & The prompts of different views.\\
   \hline
   $f(\cdot), h(\cdot), dec(\cdot)$ & The encoder, projector, and decoder.\\
    \bottomrule
\end{tabular}
}
  \vspace{-5pt}
  
\end{table}

\begin{figure*}
    \centering
    \includegraphics[width=0.95\textwidth]{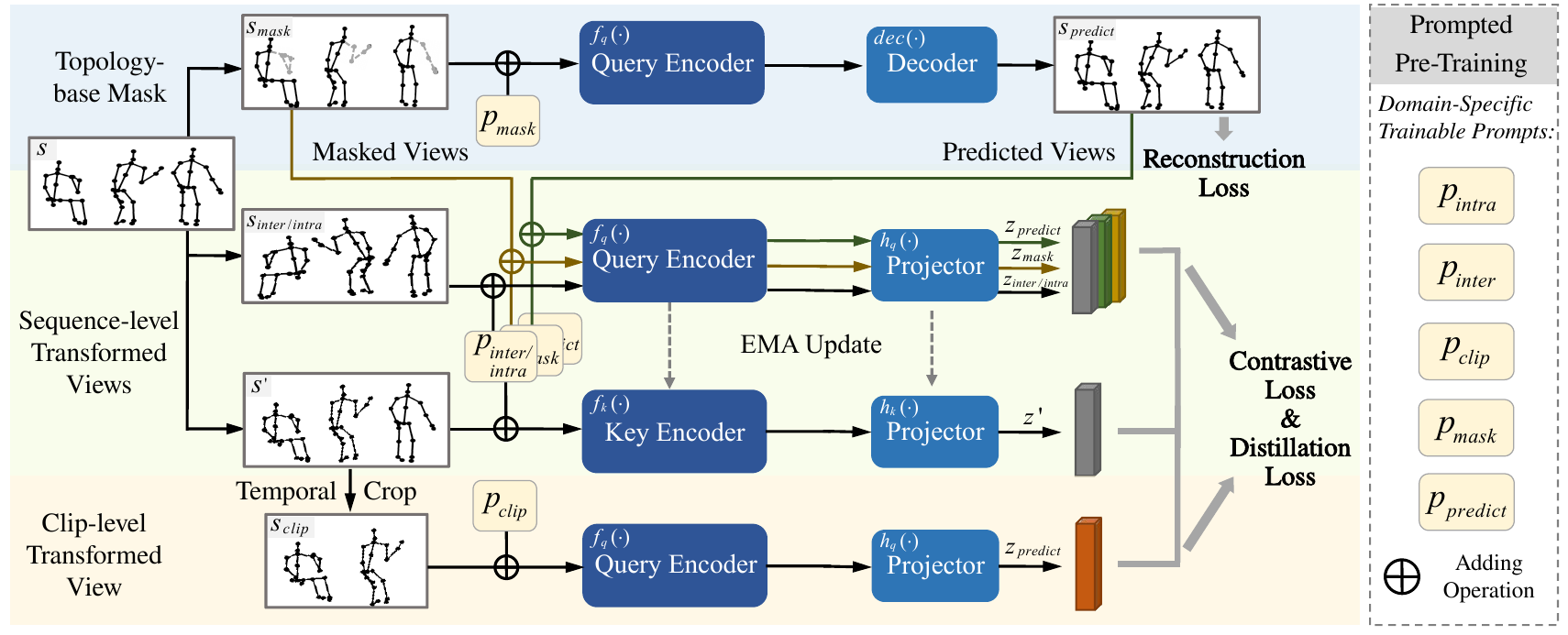}
   \caption{
    {The overview of the proposed method. We integrate masked skeleton modeling and contrastive learning paradigms to learn the joint- (\textit{blue} part), clip- (\textit{yellow} part), and sequence-level (\textit{green} part) representations. Meanwhile, a series of domain-specific prompts are introduced as training guidance. 
   }}
  \label{fig:pipeline}
\end{figure*}

\revise{Specifically, our method is based on our previous work~\cite{zhang2023prompted}, which combines contrastive learning and masked skeleton modeling in a synergetic manner. 
Based on this, we further extend and strengthen the design with the key idea, \ie, versatile action representation learning. In the following, we first introduce the contrastive learning (Sec.~\ref{sec:cl}) and MSM (Sec.~\ref{sec:mp}) scheme involved in baseline method as well as their collaboration design (Sec.~\ref{sec:syn}). Based on this, we introduce our new improvements in PCM$^{3}$++ compared to~\cite{zhang2023prompted}. Concretely, an effective clip-level contrastive learning method is designed in Sec.~\ref{sec:cl}.(3), which significantly boosts the capacity of short-term motion modeling as illustrated in Fig.~\ref{fig:pipeline}. Besides, a post-distillation strategy is developed in Sec.~\ref{sec:dis}.(2) to further boost the representation compactness. These advancements further improve the representation generalization for various downstream tasks. For clarity, the main notations are summarized in Table~\ref{tab:notation}.
}


\subsection{Skeleton Contrastive Learning}\label{sec:cl}

Our pipeline for contrastive learning follows MoCo V2~\cite{he2020momentum}. For a positive pair $(z, z^{\prime})$, the model optimizes the InfoNCE objective defined in Eq.~(\ref{eq:infonce}) along with a distillation objective in Eq.~(\ref{eq:distill}) as regularization:
\begin{equation}
    \centering
    \label{eq:info_with_distill}
    \mathcal{L}_{CL}(z, z^{\prime}) = \mathcal{L}_{Info}(z, z^{\prime}) + \mathcal{L}_{KD}(z, z^{\prime}).
\end{equation}
\revise{The intuition behind this is to combine the one-hot hard labels in InfoNCE to explicitly cluster features while avoiding collapse, and the soft labels in the distillation objective, \ie, the similarity scores, to achieve fine-grained relational knowledge learning, which also alleviates the false negative problem as discussed in Sec.~\ref{sec:contrastive_learning}.(1).
}

To boost the consistency learning, we propose a series of effective spatial-temporal augmentations to construct diverse positive pairs, which are introduced as follows.

\vspace{0.5em}

\noindent\textbf{1) Spatial Intra- and Inter- Skeleton Contrastive Learning.}
For intra-skeleton augmentations, we utilize \textit{Temporal Crop-Resize, Shear,} and \textit{Joint Jittering} to construct the positive pair $(s_{intra}, s^{\prime})$. Then, we obtain the corresponding representations $(z_{intra}, z^{\prime})$ via the query/key encoder {$f_q(\cdot)/f_k(\cdot)$} and embedding projector $h_q(\cdot)/h_k(\cdot)$, respectively. 

Meanwhile, \textit{Mixing} augmentations are adopted to construct inter-transformed views. Specifically, given two skeleton sequences $s_1, s_2$, we obtain the mixed skeleton data $s_{inter}$, \eg, by \textit{Mixup}.
Subsequently, we can obtain the embeddings corresponding to the mixed data by $z_{inter} = h_q\circ f_q(s_{inter})$. The optimized objective is $\mathcal{L}_{CL}(z_{inter}, z_{inter}^{\prime})$ in Eq.~(\ref{eq:info_with_distill}),
where $z_{inter}^{\prime} = (1-\lambda) (h_{k}\circ f_k(s_1)) + \lambda (h_{k}\circ f_k(s_2))$.

\vspace{0.5em}
\noindent\textbf{2) Temporal Asymmetric Clip Contrastive Learning.} The above designs only consider the sequence-level consistency modeling, \ie, the augmented data often contains sufficient temporal motion information. However, it ignores the \textit{short-term} motion modeling, which is necessitated for downstream tasks of dense prediction such as action detection. On the other hand, the consistency learning for short-term motions, \eg, clips, which can be viewed as challenging temporal augmented views, can also benefit the whole representation quality learned by model. To this end, we present the clip-level contrastive learning design. 

To sample a motion clip, we can apply the \textit{Temporal Crop} transformation, and the obtained clip is much shorter ($10\%\sim40\%$) than the original sequence.
However, the clip only contains partial motion information and directly aligning the semantic embeddings of two clips is difficult for model, leading to the unstable training. Therefore, we introduce an asymmetric design, which only feeds the short clip into query branch, while the key branch takes the pre-crop original sequence as an anchor. Meanwhile, we keep the two augmented data possessing the same spatial views to decouple the temporal variation.
Specifically, we first apply the intra-augmentations to a sample $s$ to generate the $s_{ach}$. Then we further utilize the \textit{Temporal Crop} to sample a continuous clip $s_{clip}$ from $s_{ach}$. It means that, $s_{ach}$ is subsuming $s_{clip}$ in temporal dimension while sharing exactly the same spatial transformations, to provide a precise and feasible target for clip-level consistency learning. 
Similarly, the model takes the positive pair $(z_{clip}, z_{ach})$ for $\mathcal{L}_{CL}$ optimization in Eq.~(\ref{eq:info_with_distill}),
where $z_{clip} = h_q \circ f_q (s_{clip})$. In implementation, we just mask $z_{ach} = z^{\prime}$ for efficiency. 


\subsection{Masked Skeleton Prediction}\label{sec:mp}
To model the joint-level representations, we utilize the masked skeleton modeling task with a segment-wise masking strategy at body part level. 
To predict the masked regions from masked skeleton $s_{mask}$, we employ a decoder $dec(\cdot)$ taking the representations from the encoder as input. The MSE loss between original data $s$ and predicted data $s_{predict}$ by decoder is optimized in the masked regions:
\begin{equation}
\label{eq:mask}
\mathcal{L}_{Mask} = \mathbb{E}\left(||(s - dec\circ f_q(s_{mask}))\odot (\mathbf{1}-M) ||_{2}\right),
\end{equation}
where 
$M$ is the binary mask and $\mathbf{1}$ is an all-one matrix with the same shape as $M$.

\subsection{On the Connection of Contrastive Learning and Masked Prediction}\label{sec:syn}
As discussed in the work~\cite{zhang2023prompted}, simply combining these two tasks can be sub-optimal due to the inherent gap of their feature modeling mechanisms~\cite{qi2023contrast,gui2023survey}.
To this end, we explore the potential synergy for exploitation between them.

\vspace{0.5em}

\noindent\textbf{1) Novel Positive Pairs as Connection.} We utilize special data views during the masked prediction training to provide more diverse positive samples for contrastive learning. 
First, the masked skeleton view $s_{mask}$ naturally simulates the occlusion for skeletons, serving as challenging positives. Meanwhile, we also take the predicted skeleton view $s_{predict}$ output by decoder $dec(\cdot)$ as positive samples. It contains the inherent uncertainty and diversity brought by continuous training of the model, which contributes to encoding more diverse movement patterns. On the other hand, the semantic consistency of the output skeleton with respect to the model itself is encouraged, \ie, the predicted view can also be perceived well by the encoder, connecting the low-level reconstruction with the high-level semantic modeling.

In a nutshell, we utilize the masked view $s_{mask}$ and the predicted view $s_{predict}$ as positives. Together with the manually constructed positive pairs in Sec.~\ref{sec:cl}, we present all positive (embedding) pairs $\{(z_{q}, z_{k})\}$ as follows:
\begin{equation}
\begin{aligned}
\label{eq:pos}
\{(z_{q}, z_{k})\} = 
\{(&z_{intra}, z^{\prime}), (z_{inter}, z^{\prime}_{inter}), (z_{clip}, z^{\prime}), \\(&z_{mask}, z^{\prime}), (z_{predict}, z^{\prime}) \}.
\end{aligned}
\end{equation}
They are fed into the encoders to optimize the $\mathcal{L}_{Info}$ and $\mathcal{L}_{KD}$ in Eq.~\ref{eq:info_with_distill}.



\vspace{0.5em}

\noindent\textbf{2) High-Level Semantic Guidance.}
Note that we feed the predicted view into the contrastive learning pipeline. The gradients of $s_{predict}$ from the contrastive learning branch are propagated to update the reconstructed decoder $dec(\cdot)$. It provides the high-level semantic guidance for the skeleton prediction together with the joint-level MSE loss in Eq.~(\ref{eq:mask}), leading to better masked prediction learning and higher quality of $s_{predict}$ as positive samples.

\revise{With these synergetic designs, the masked skeleton modeling provides novel positives to the contrastive learning, boosting the model generalization capacity. On the other hand, the supervision signals as gradients of contrastive learning branch also assists the masked prediction branch with the high-level semantic guidance provided by inter-sample contrast. These designs alleviate the gap of feature modeling mechanism of these two tasks, which limits the performance of previous works that directly combine them, and hence yield better representation quality.}

\subsection{The Whole Training Strategy}\label{sec:dis}
Overall, the model jointly optimizes the contrastive learning and masked skeleton modeling. Here we present the overall objective and the training strategy. For clarity, we utilize $\mathcal{L}_{Info}^{all}$ and $\mathcal{L}_{KD}^{all}$ to represent the sum of the losses for each positive pair defined in Eq.~(\ref{eq:pos}). It can be formulated as:
\begin{equation*}
    \mathcal{L}_{Info}^{all} = \sum_{(z_{q}, z_{k})}\mathcal{L}_{Info}^{query}, \ \ 
    \mathcal{L}_{KD}^{all} = \sum_{(z_{q}, z_{k})}\mathcal{L}_{KD}^{query}.
\end{equation*}
$\mathcal{L}_{Info}^{query}$ and $\mathcal{L}_{KD}^{query}$ are the respective terms for the specific $query$ view summarized in Eq.~(\ref{eq:pos}).
During training, the following objective is applied to the whole model:
\begin{equation}
\label{eq:overall_loss}
	\mathcal{L} = \mathcal{L}_{Info}^{all} + \lambda_{m}\mathcal{L}_{Mask} + \lambda_{kd}\mathcal{L}_{KD}^{all},
\end{equation}
where $\lambda_{m}$ and $\lambda_{kd}$ are the loss weight. Based on this, we further propose two advanced training strategies to boost the representation learning.

\vspace{0.5em}
\noindent\textbf{1) Prompted Multi-Task Pre-Training.} Considering the difficulty of encoding different data views simultaneously, we employ prompt-based guidance to assist the model to learn from different data views explicitly.
Specifically, we attend to a series of \textit{domain-specific prompts} for different augmented views, \ie, $p_{intra}, p_{inter}, p_{clip}, p_{mask}$, and $p_{predict}$. Their dimension equals to the spatial size of the skeleton data. Then, these domain-specific prompts are added to the corresponding skeleton ($s_*$ means $*$ view of $s$):
\begin{equation}
	s_{*} = s_{*} + p_{*}.
\end{equation}
These decorated skeletons are fed into the query/key encoder for self-supervised pre-training, providing the training guidance and achieving better representations. 
\vspace{0.5em}

\noindent\textbf{2) Post-Distillation Refinement.} To further improve the representation quality, we introduce a post-distillation strategy as an optional refinement process. After obtaining a good prior feature space by Eq.~(\ref{eq:overall_loss}), we directly remove the one-hot label constraint in InfoNCE objective $\mathcal{L}_{Info}^{all}$, \ie,
\begin{equation}
\label{eq:post_loss}
	\mathcal{L} = \lambda_{m}\mathcal{L}_{Mask} + \lambda_{kd}\mathcal{L}_{KD}^{all},
\end{equation}
where we only apply distillation term that assigns the attraction weights adaptively according to the similarity. This can be seen as a more explicit feature clustering process to obtain a more compact representation space by alleviating the \textit{false negative} problem. Note this single distillation objective would not cause model collapse due to the well initialized representation space learned in previous pre-training stage.

\subsection{Discussion}
\revise{We summarize our method as a direct response to the challenges mentioned in previous literature review. We target at the versatile action representation learning, adopting our previous work~\cite{zhang2023prompted} as the baseline for its effectiveness to learn multi-granularity features. Building on this, we further integrate the clip-level contrastive learning with it to present a more comprehensive solution with joint, clip, and sequence level modeling.}

\revise{As a response to previously discussed challenges, 1) we carefully design the spatial-temporal augmentations involving intra- and inter- data samples to present a strong contrastive learning baseline, which eases the efforts of subsequent works. 2) Further, the proposed multi-granularity representation modeling can effectively enhance the feature diversity, alleviating the dimensional collapse implicitly according to~\cite{zhang2022mask}. 3) Meanwhile, for the false negative problem, we employ a different perspective, \ie, post-training to deal with it. In the following sections, we will present the quantitative analysis on these designs.}

\section{Benchmark Skeleton SSL Methods}
\label{sec:experiment}
\label{sec:survey:evaluation}
{We provide a comprehensive benchmark of existing methods, along with the skeleton datasets, backbone architectures, and downstream tasks in this section. Meanwhile, the evaluation of proposed PCM$^{3}$++ is reported.}

We first introduce the following datasets, which are widely used in skeleton-based SSL evaluation. The summary of more datasets for skeleton action understanding can be found in Supp.

\noindent\textbf{1) NTU RGB+D 60 Dataset (NTU 60)}~\cite{shahroudy2016ntu} is the most popular skeleton dataset. There are 56,578 videos with 25 joints for a human, captured by three Microsoft Kinect v2 cameras. The skeleton sequences are divided into 60 action categories, performed by 40 volunteers.
Two evaluation protocols are recommended:  a) Cross-Subject (xsub): the data for training are collected from 20 subjects, while the other 20 subjects are for testing. b) Cross-View (xview):  the training set consists of front and two side views of the action performers, while testing set includes the left and right 45 degree views. 

\noindent\textbf{2) NTU RGB+D 120 Dataset (NTU 120)}~\cite{liu2019ntu} is an extension to NTU 60 dataset. There are 114,480 videos collected with 120 action categories, performed by 106 subjects. Meanwhile, 32 collection setups with respect to the location and background are used to build the dataset. Two recommended protocols are presented: a) Cross-Subject (xsub): the data for training are collected from 53 subjects, while the testing data are from the other 53 subjects. b) Cross-Setup (xset): the training data uses even setup IDs, while testing data are odd setup IDs.

\noindent\textbf{3) PKU Multi-Modality Dataset (PKUMMD)}~\cite{liu2020pku}{ is another large-scale benchmark with available skeleton data. Two subsets, Part I and Part II, are provided. 
PKUMMD Part I contains 1,076 long video sequences, with 20 action labels per video on average, and $\sim$20,000 instances are included in 51 action categories after trimming. Part II contains 2000 short video sequences with approximately seven instances each, focusing on the short-margin action detection task. It is more challenging due to the data noise and view variation. 
}

\begin{table*}[!h]

\caption{Comparison of skeleton SSL works. P, G and C represents context-based (Pseudo-label), Generative, and Contrastive learning methodologies. We report the best accuracy in the original paper. *s means the fusion results of * streams, and the single joint stream is adopted by default.
}
\label{tab:main_comp}
\center
\small
\color{black}
\resizebox{\linewidth}{!}{%
\setlength{\tabcolsep}{3.0mm}{
\begin{tabular}{l | l | c | c | c | c c| cc}
\toprule
\multirow{2}{*}{{Method}} & \multirow{2}{*}{{Publish}} & \multirow{2}{*}{{Backbone}} & {{Feature}} & {Pretext Task} &  \multicolumn{2}{c|}{NTU 60 (\%)} & \multicolumn{2}{c}{NTU 120 (\%)} \\
 & & &Dimension &P $\|$ G $\|$ C &xsub &xview &xsub &xset\\
\toprule
\multicolumn{9}{c}{\textit{Linear Evaluation Protocol} (Arranged by Backbone Model and Publish Year)} \\
LongT GAN~\cite{zheng2018unsupervised} & AAAI 2018 & GRU & 800 & $\circ$ $\|$ $\bullet$ $\|$ $\circ$ & 39.1 & 52.1 & 35.6 & 39.7\\
P\&C~\cite{su2020predict} & CVPR 2020 & GRU & 1024$\times$2 &$\circ$ $\|$ $\bullet$ $\|$ $\circ$& 50.7 & 76.1 & 41.1 & 44.1 \\
2s-SeBiReNet~\cite{nie2020unsupervised} & ECCV 2020 & GRU &32  & $\circ$ $\|$ $\bullet$ $\|$ $\bullet$& - & 79.7 & - & 69.3 \\
MS$^{\rm 2}$L~\cite{lin2020ms2l} & ACM MM 2020 & GRU & 600 & $\bullet$ $\|$ $\bullet$ $\|$ $\bullet$& 52.6 & - & - & -  \\
PCRP~\cite{xu2021prototypical} & TMM 2021 & GRU &1024  & $\circ$ $\|$ $\bullet$ $\|$ $\bullet$ & 54.9&63.4&43.0 &44.6\\
AS-CAL~\cite{rao2021augmented} & Info. Sciences 2021 & LSTM & 256 & $\circ$ $\|$ $\circ$ $\|$ $\bullet$& 58.5 & 64.8 & - & - \\ 
ISC~\cite{thoker2021skeleton} & ACM MM 2021 & GRU+GCN & 1024$\times$2 & $\circ$ $\|$ $\circ$ $\|$ $\bullet$ & 76.3 & 85.2 & 67.1 & 67.9 \\
CRRL~\cite{wang2022contrast} & TIP 2022 & GRU & 300 & $\circ$ $\|$ $\bullet$ $\|$ $\bullet$& 67.6 & 73.8 & 57.0 & 56.2\\
3s-CMD~\cite{mao2022cmd} & ECCV 2022 & GRU & 1024$\times$2 & $\circ$ $\|$ $\circ$ $\|$ $\bullet$& 84.1 & 90.9 & 74.7 & 76.1 \\  
3s-CSTCN~\cite{wang2023learning} & TMM 2023 & GRU & 1024$\times$2 & $\circ$ $\|$ $\circ$ $\|$ $\bullet$& 85.8 & 92.0 & 77.5 & 78.5 \\ 
3s-HiCo~\cite{chen2022hierarchically} & AAAI 2023 & GRU & 512$\times$8 & $\circ$ $\|$ $\circ$ $\|$ $\bullet$& 82.6 & 90.8 & 75.9 & 77.3\\   
HaLP~\cite{shah2023halp} & CVPR 2023 & GRU & 1024$\times$2 & $\circ$ $\|$ $\circ$ $\|$ $\bullet$& 79.7 & 86.8 & 71.1 & 72.2  \\     
3s-PCM$^{\rm 3}$~\cite{zhang2023prompted} & ACM MM 2023 & GRU & 1024$\times$2 & $\circ$ $\|$ $\bullet$ $\|$ $\bullet$& 87.4 & 93.1 & 80.0 & 81.2 \\
3s-Eq-Contrast~\cite{lin2024mutual} & TIP 2024 &GRU & 1024$\times$2 & $\bullet$ $\|$ $\circ$ $\|$ $\bullet$&87.0& 92.9 & 79.4 & 81.2\\
\textbf{3s-PCM$^{\rm 3}$++} & - (This Paper) & GRU & 1024$\times$2 & $\circ$ $\|$ $\bullet$ $\|$ $\bullet$& 88.1 & 93.5 & 80.3 & 81.6 \\
\midrule
     
3s-CrosSCLR~\cite{li20213d} & CVPR 2021 & GCN & 256 & $\circ$ $\|$ $\circ$ $\|$ $\bullet$& 77.8 & 83.4 & 67.9 & 66.7 \\
4s-MG-AL~\cite{yang2022motion} & TCSVT 2022 & GCN & - & $\bullet$ $\|$ $\circ$ $\|$ $\circ$  & 64.7 & 68.0 & 46.2 & 49.5  \\
3s-AimCLR~\cite{guo2022aimclr} & AAAI 2022 & GCN & 256 & $\circ$ $\|$ $\circ$ $\|$ $\bullet$& 78.9 & 83.8 & 68.2 & 68.8 \\
3s-CPM~\cite{zhang2022cpm} & ECCV 2022 & GCN & 256 & $\circ$ $\|$ $\circ$ $\|$ $\bullet$& 83.2 & 87.0 & 73.0 & 74.0 \\
Chen et.al~\cite{chen2023self} & TIP 2023 & GCN & 256 & $\circ$ $\|$ $\circ$ $\|$ $\bullet$& 78.9 & 82.3 & 68.4 & 67.3  \\
3s-HiCLR~\cite{zhang2022hiclr} & AAAI 2023 & GCN & 256 & $\circ$ $\|$ $\circ$ $\|$ $\bullet$& 80.4 & 85.5 & 70.0 & 70.4  \\
3s-PSTL~\cite{zhou2023self} & AAAI 2023 & GCN & 256 & $\circ$ $\|$ $\circ$ $\|$ $\bullet$& 79.1 & 82.6 & 69.2 & 70.3 \\
3s-SkeAttnCLR~\cite{hua2023part} & IJCAI 2023 & GCN & 256 & $\circ$ $\|$ $\circ$ $\|$ $\bullet$& 82.0 & 86.5 & 77.1 & 80.0  \\
3s-HYSP~\cite{francohyperbolic} & ICLR 2023 & GCN & 256 & $\circ$ $\|$ $\circ$ $\|$ $\bullet$& 79.1 & 85.2 & 64.5 & 67.3  \\   
3s-ActCLR~\cite{lin2023actionlet} & CVPR 2023 & GCN & 256 & $\circ$ $\|$ $\circ$ $\|$ $\bullet$& 84.3 & 88.8 & 74.3 & 75.7 \\ 
3s-RVTCLR+~\cite{zhu2023modeling} & ICCV 2023 & GCN & 256 & $\circ$ $\|$ $\circ$ $\|$ $\bullet$& 79.7 & 84.6 & 68.0 & 68.9 \\
2s-ViA~\cite{yang2022via} & IJCV 2024 & GCN  &256	& $\circ$ $\|$ $\bullet$ $\|$ $\bullet$&	78.1 & 85.8 & 69.2 & 66.9	\\		

\midrule

H-Transformer~\cite{cheng2021hierarchical} & ICME 2021 & Transformer & 2048 &$\bullet$ $\|$ $\circ$ $\|$ $\circ$& 69.3 & 72.8 & - & - \\
GL-Transformer~\cite{kim2022global} & ECCV 2022 & Transformer & 48$\times$25 & $\bullet$ $\|$ $\circ$ $\|$ $\circ$ & 76.3 & 83.8 & 66.0 & 68.7 \\
MAMP~\cite{mao2023masked} & ICCV 2023 & Transformer & 256 &$\circ$ $\|$ $\bullet$ $\|$ $\circ$& 84.9 & 89.1 & 78.6 & 79.1 \\
3s-UmURL~\cite{sun2023unified} & ACM MM 2023 & Transformer & 2048 & $\circ$ $\|$ $\circ$ $\|$ $\bullet$& 84.4 & 91.4 & 75.9 & 77.2 \\
MacDiff~\cite{wu2025macdiff} & ECCV 2024 & Transformer& 256$\times$25 & $\circ$ $\|$ $\bullet$ $\|$ $\circ$ & 86.4 & 91.0 & 79.4 &80.2 \\
IGM~\cite{lin2025idempotent} & ECCV 2024 & Transformer& 256$\times$25& $\circ$ $\|$ $\bullet$ $\|$ $\bullet$ &86.2 &91.2 &80.0 &81.4\\
3s-USDRL~\cite{weng2025usdrl} & AAAI 2025 & Transformer & 4096 & $\circ$ $\|$ $\circ$ $\|$ $\bullet$ & 87.1 & 93.2 &79.3 & 80.6 \\

\midrule

AE-L~\cite{paoletti2022unsupervised}&BMVC 2021	&CNN 	&256	&$\bullet$ $\|$ $\bullet$ $\|$ $\circ$  &69.9 &	85.4	 &59.1 &	62.4 \\
3s-Colorization~\cite{yang2021skeleton} & ICCV 2021 & DGCNN & 1024 &$\circ$ $\|$ $\bullet$ $\|$ $\circ$& 75.2 & 83.1 & 64.3 & 67.5  \\
3s-Masked Colorization~\cite{yang2023self} & TPAMI 2023 & DGCNN & 1024 &$\circ$ $\|$ $\bullet$ $\|$ $\circ$& 79.1 & 87.2 & 69.2 & 70.8 \\

\midrule
\midrule
\multicolumn{9}{c}{\textit{Fully Fine-tuning Protocol} (Arranged by Publish Year)}\\
MCC~\cite{su2021self}	& ICCV 2021	& GCN &256& $\circ$ $\|$ $\bullet$ $\|$ $\bullet$& 83.0 & 89.7 & 77.0	 & 77.8  \\
3s-Hi-TRS~\cite{chen2022hierarchically} &ECCV 2022  &Transformer&512 & $\bullet$ $\|$ $\bullet$ $\|$ $\bullet$ &90.0	&95.7	&85.3	&87.4 \\
3s-Masked Colorization~\cite{yang2023self}	& TPAMI 2023  &DGCNN&1024& $\circ$ $\|$ $\bullet$ $\|$ $\circ$&	89.1	&95.9&	81.2&	82.4 \\
MAMP~\cite{mao2023masked} &ICCV 2023& Transformer &256$\times$25 &$\circ$ $\|$ $\bullet$ $\|$ $\circ$  &93.1	&97.5 &90.0	&91.3\\
MotionBERT~\cite{zhu2023motionbert}	& ICCV 2023	&Transformer &512  &$\circ$ $\|$ $\bullet$ $\|$ $\circ$ &93.0&	97.2&	-&	- \\
SSL~\cite{yan2023skeletonmae}	&ICCV 2023&	GIN& - &$\circ$ $\|$ $\bullet$ $\|$ $\circ$	&92.8&	96.5&	84.8&	85.7 \\
3s-PCM$^{\rm 3}$++ & - (This paper) & GRU & 1024$\times$2 & $\circ$ $\|$ $\bullet$ $\|$ $\bullet$& 88.5 & 93.9 & - & - \\

\bottomrule
\end{tabular}
}
}
\vspace{-5pt}
\label{tab:methods}
\end{table*}

\begin{table*}[!h]

\caption{\revise{Benchmark on multiple downstream tasks. *UmURL adopts hard-coded input temporal length and the performance drops significantly when handling the fine-grained temporal understanding task. $\dag$ indicates more input frames are used.}
}
\label{tab:other_tasks}
\center
\color{black}
\resizebox{\linewidth}{!}{%
\setlength{\tabcolsep}{3.0mm}{
\begin{tabular}{l | cccc | cc | cc | cc }
\toprule
\multirow{3.5}{*}{{Method}} & \multicolumn{4}{c|}{Retrieval} & \multicolumn{2}{c|}{Occluded} & \multicolumn{2}{c|}{Detection} & \multicolumn{2}{c}{Few Shot} \\
\cmidrule(r){2-5} \cmidrule(r){6-7} \cmidrule(r){8-9} \cmidrule(r){10-11}
& \multicolumn{2}{c}{NTU 60} & \multicolumn{2}{c|}{NTU 120} & \multicolumn{2}{c|}{NTU 60 xview} & \multicolumn{2}{c|}{PKUMMD} & \multicolumn{2}{c}{NTU 120 xset} \\
&xsub &xview &xsub &xset & Spatial &Temporal & mAP@0.1 & mAP@0.5 & (5,1) & (5,5) \\
\toprule
\multicolumn{9}{l}{\textit{GRU-based methods}} \\
MoCo-GRU~\cite{he2020momentum} & 62.5 &82.0 &52.2 &55.6 &72.6 & 74.8 & 68.2 & 63.4&58.5 & 78.7 \\
ISC~\cite{thoker2021skeleton} &62.5	&82.6 &50.6 &	52.3 &70.6 & 76.8 &64.6 & 58.7 &62.3 & 81.3 \\
CRRL~\cite{qi2023contrast} &60.7	&75.2	&-	&- & 61.4 & 66.2 & 57.6 &52.1&58.3&77.1\\
CMD~\cite{mao2022cmd} & 70.6	&85.4  	&58.3	&60.9 &72.7&79.5&73.7 & 68.4 & 64.9 & 82.8\\
HaLP~\cite{shah2023halp}&65.8 &83.6 &55.8 &59.0 & 72.0 & 79.8 & 71.1 & 64.9 & 64.2 & 82.9\\
HiCo~\cite{hico2023} & 67.9 & 84.4 & 55.9 &58.7 & 72.1 & 76.7 &51.8 & 45.8 & 64.6 & 81.7\\
PCM$^3$~\cite{zhang2023prompted}& 73.7 &88.8 &63.1 &66.8 & 87.0 & 86.1&73.3&68.2 & 65.2 & 82.9\\
PCM$^3$++ &{75.4} &{89.4} &{64.5} &{67.1} & 88.0 & 85.8&75.5 & 69.8 & 66.7 &84.8\\

\midrule
\multicolumn{9}{l}{\textit{GCN-based methods}} \\
AimCLR~\cite{guo2022aimclr} &58.8 &67.9 &43.7 &42.4 &48.5 &54.7 & 43.9 & 35.1 & 66.7 & 83.8 \\
HiCLR~\cite{zhang2022hiclr} &60.6 &73.1 &46.0 &46.0 &46.8 &53.3 & 36.7 & 29.2 &65.0 &82.2\\
SkeAttnCLR~\cite{hua2023part} & 55.3 &60.2 &- &- &47.4 &52.9 & 48.5 & 41.7 & 62.1 & 76.9  \\
ActCLR~\cite{lin2023actionlet} & 69.6 &76.1 &51.4 &54.9 &48.3  &53.3 & 41.9& 35.1 & 68.0 & 84.3 \\

\midrule
\multicolumn{9}{l}{\textit{Transformer-based methods}} \\
MAMP$\dag$~\cite{mao2023masked}& 62.0 & 70.0 &- & - &56.5 &81.0 &78.8 &72.4 &56.2 & 76.3\\
UmURL~\cite{sun2023unified} &72.0 & 88.9 & 59.5 & 62.2 &71.1 &77.4 &43.4* &38.3* &64.9 &77.2  \\

\bottomrule
\end{tabular}
}
}
\vspace{-5pt}
\label{tab:methods}
\end{table*}


\subsection{Model Backbones}
\label{sec:backbone}
Different backbones are studied in previous skeleton SSL works, \ie, RNN-based, GCN-based, and Transformer-based models. RNNs treat the skeleton sequences as the temporal series and model temporal dependencies. However, it ignores the spatial structures of skeleton. Inspired by the natural topology structure of the human body, GCNs are widely explored to model spatial-temporal relationships. Recently, Transformer has been utilized to capture the long-temporal dependencies and has demonstrated remarkable results, owing to the attention mechanism. 

Besides, some SSL works turn to other model backbones. Convolutional neural networks (CNNs) are utilized to process the skeleton sequence as a pseudo-2D image. Meanwhile, skeletons can also be treated as point clouds, and hence the Dynamic Graph CNN (DGCNN)~\cite{wang2019dynamic} is also explored as the feature extractor.

\subsection{Downstream Tasks}
\label{sec:task}
\noindent \textbf{1) Action Recognition} is the most common SSL evaluation task. Typically, two evaluation protocols are widely adopted. The first is the linear evaluation protocol, where a linear layer is added with the pre-trained model fixed. The other is fine-tuning protocol where the whole model is trained including the subsequent linear layer. Top-k accuracy metric is adopted in this task.

\noindent \textbf{2) Action Retrieval} aims to find the skeleton sequences that are similar or near-duplicates of a given query sequence. The metric precision, which is the proportion of retrieved relevant skeleton sequences in all retrieved entries, is reported. 

\noindent \textbf{3) Action Detection} is a task that detects the start and end time of actions in an untrimmed skeleton sequence as well as its corresponding action label. It can also be referred to as the \textit{Temporal Action Localization} or \textit{Action Segmentation}. Following the previous works~\cite{luo2018graph,liu2020pku}, the Average Precision (mAP) at different temporal Intersection over Union (tIoU) thresholds between the predicted and the ground truth intervals is utilized as the metric. 

\noindent \textbf{4) Occluded Action Recognition} focuses on the action understanding with occlusion, which is prevalent in human activity. Typically, the skeleton joints with low confidence or known to be occluded are set to zeros. The model is constrained to predict the correct label from corrupted skeleton sequences with top-k accuracy as the metric.

\begin{figure}[t]
    \centering
    \includegraphics[width=0.49\textwidth]{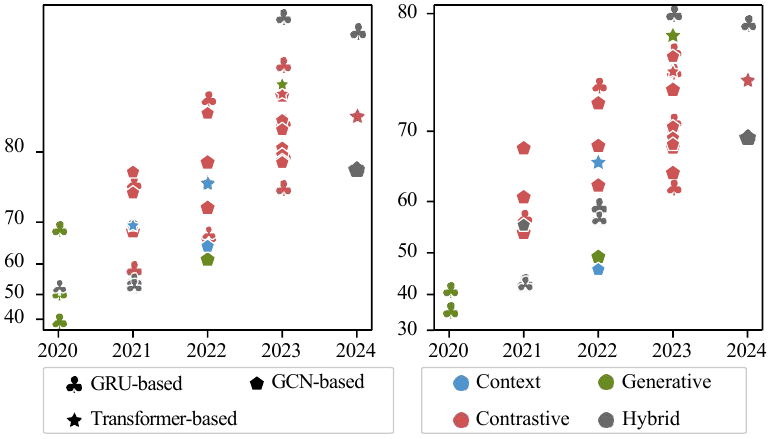}
   \caption{
   Action recognition performance of model over time for different SSL methodologies and different backbones. Left: NTU 60, Right: NTU 120, using cross-subject linear evaluation protocol.
   }
   \vspace{-10pt}
  \label{fig:scatter}
\end{figure}

\subsection{Implementations Details}
For the proposed method, we follow the implementation and experiment settings of previous works~\cite{zhang2023prompted,mao2022cmd}. For the existing methods, we give priority to the reported results in their original paper. Meanwhile, we reproduce some results to keep the fairness of the comparison and implementation. Details can be found in Supp.

\subsection{\revise{A Multi-Task Benchmark for Skeleton-based SSL}}
In this section, we present our multi-task benchmark for skeleton-based SSL methods, which is the first time to the best of our knowledge, to benefit the community. Considering most previous SSL works are evaluated only on action recognition, we split the benchmark into two parts for presentation, \ie, a full comparison on action recognition task and the reproduction benchmark on other tasks. 
First, we curate a comprehensive comparison of the widely adopted action recognition task in Table~\ref{tab:main_comp} with the backbone and pretext task category mark. 

{On the other hand, for other different tasks, an available benchmark is provided covering action retrieval, occluded recognition, action detection, and few-shot learning tasks. We select the recent popular works with official code released, and make huge efforts to reproduce them on the same settings to present a multi-task comparison as shown in Table~\ref{tab:other_tasks}.}
\vspace{0.2em}

\subsubsection{Comparison
on Action Recognition Task}

\noindent{\textbf{1) {Across Pretext Tasks}}.}
As shown in Fig.~\ref{fig:scatter}, context-based and generative learning pretext tasks are mainly studied in the earlier works. However, these methods usually fail to achieve a satisfactory performance under linear evaluation due to the modeling of too many low-level features and the lack of effective design. Recently, MAMP~\cite{mao2023masked} adopts MSM with large mask ratios and motion-aware reconstruction targets, obtaining a highly discriminative semantic feature space for both linear evaluation and fully fine-tuning protocol. The subsequent works~\cite{lin2025idempotent,wu2025macdiff} also demonstrate promising results for Transformer-based MSM. However, these methods are only reported based on the joint view, which makes cross-modal knowledge ensemble in this paradigm still an open problem.

Contrastive learning has always been a popular topic triggered by the success in the image field, \eg, SimCLR and MoCo. The model can learn a separable high-level representation space, making it dominant in learning linear representations. These works start from the exploiting the baseline algorithms~\cite{rao2021augmented,thoker2021skeleton}, and significantly boost the performance by applying stronger augmentations~\cite{guo2022aimclr,zhang2022hiclr,lin2023actionlet}, effective training strategies~\cite{zhang2022cpm,chen2023self}, cross-modal knowledge~\cite{li20213d,mao2022cmd}, achieving rapid improvement in recent three years. 

Besides, combining different pre-training paradigms~\cite{zhang2023prompted,lin2024mutual} for skeleton-based representation learning has demonstrated promising results recently. This indicates that the representations learned by different pre-training paradigms can be complementary and beneficial.

\vspace{0.2em}
\noindent \textbf{2) Across Model Backbones}.
Different model backbones often employ different pre-training tasks. For example, Transformer often consumes extensive computational resources, making it difficult for contrastive learning, which often requires encoding multiple data views. In contrast, it naturally fits the MSM schema which reduces computational overhead by masking tokens with large ratios. On the other hand, GRU and GCN models are relatively more efficient in training, often adopting the contrastive learning pretext task. It is also noted that the GRU models are often with larger feature dimensions than GCN models due to the smaller induced computational graph and less GPU memory occupancy.

Generally, different from supervised training, GRU models have achieved state-of-the-art performance under linear evaluation as shown in Fig.~\ref{fig:scatter} in SSL, and the Transformer models are also promising recently. \revise{For fully fine-tuning protocol, GRU model is less effective due to the over-fitting caused by high dimension and huge parameters. Instead, Transformer models are more popular due to the powerful attention mechanism.} Remarkably, with self-supervised pre-training, a vanilla Transformer~\cite{mao2023masked} has surpassed the GCN/Transformer with complex designs in supervised training~\cite{lee2023hierarchically,shi2020decoupled}, demonstrating the significant role for alleviating the over-fitting problem and generalization capacity improvement of SSL.

\vspace{0.2em}

\subsubsection{\revise{Comparison on Other Downstream Tasks}}

\begin{table*}[!h]

\caption{\revise{Analytics on the pre-training of different datasets for multiple downstream tasks.}
}
\label{tab:dataset_dependency}
\center
\color{black}
\resizebox{\linewidth}{!}{%
\setlength{\tabcolsep}{3.0mm}{
\begin{tabular}{l | cc| cc | cc | cc | cc }
\toprule
\multirow{3.5}{*}{{Pre-training}} & \multicolumn{2}{c|}{Recognition} & \multicolumn{2}{c|}{Retrieval} & \multicolumn{2}{c|}{Occluded} & \multicolumn{2}{c|}{Detection} & \multicolumn{2}{c}{Few Shot} \\
\cmidrule(r){2-5} \cmidrule(r){6-7} \cmidrule(r){8-9} \cmidrule(r){10-11}
& NTU 60 & NTU 120 & NTU 60 & NTU 120 & \multicolumn{2}{c|}{NTU 60 xview} & \multicolumn{2}{c|}{PKUMMD} & \multicolumn{2}{c}{NTU 120 xset} \\
&xsub &xsub &xsub &xsub & Spatial &Temporal & mAP@0.1 & mAP@0.5 & (5,1) & (5,5) \\
\toprule

NTU 60 &84.8 &73.5 &75.4 &58.8 &88.0  & 85.8&75.5 & 69.8 & 66.7 &84.8\\
NTU 120 &85.1  &76.7  &75.3  & {64.5} & 87.7&88.2 &75.3 &69.4 &69.6 &87.2\\
PKUMMD &77.6 &67.5  &65.1 &52.9 &78.0 & 79.3& 70.7 & 65.7 &58.9&78.9\\
NTU 120+PKUMMD &85.8 &77.0 &76.2&64.5 &88.4 & 88.0  &76.2 &70.5 &69.4 &87.0\\
\bottomrule
\end{tabular}
}
}
\vspace{-5pt}
\label{tab:methods}
\end{table*}

\vspace{0.1em}
\noindent \textbf{1) Experiment Setting.} We introduce the implementations of each task first and more details can be found in Supp.

\textit{Skeleton-based Action Retrieval}. Following previous work~\cite{su2020predict}, a K-nearest neighbors (KNN) classifier ($k$=1) is adopted to retrieve the nearest training sample for each testing data in the representation space. All models are first pre-trained and then evaluated on the target dataset. The precision is reported as accuracy.

\textit{Action Recognition with Occlusion.} We evaluate the transfer ability of representations learned from the clean dataset to the action recognition with occluded data. We adopt the linear evaluation protocol. Following~\cite{song2019richly}, we construct a synthetic occluded dataset on both spatial and temporal dimension. For spatial dimension, different body parts are randomly masked. For temporal dimension, we set a random block of frames to zeros. The testing set is constructed by the same masks across different methods, with a masking ratio of [0.3, 0.7]. 

\textit{Skeleton-based Action Detection.} We follow the previous works~\cite{luo2018graph,liu2020pku}, and evaluate the short-term motion modeling capacity by action detection task. We train the attached linear classifier to predict frame-level categories to produce the final proposal. The encoder is pre-trained on NTU 60 dataset and then we fix it, transferring the learned representations to untrimmed PKUMMD Part I dataset. The mAP of different actions is adopted as metric with different tIoUs. 

\textit{Unsupervised Few-shot Learning.} We evaluate the performance of SSL as few-shot learners follow the previous work~\cite{lu2022self}. Specifically, the model is first pre-trained on NTU 60 dataset. Then, a simple classifier, \eg, Support Vector Machine (SVM), is fitted on the output features by the pre-trained encoder of the support data set. Finally, the adapted classifier along with the encoder is utilized to infer the query samples. We select 20 new categories in NTU 120\footnote{Specifically, the classes (index from 0) 60, 61, 66, 69, 72, 78, 79, 80, 84, 90, 91, 95, 96, 98, 99, 100, 102, 106, 108, 111, 113 114, 115 are selected.} which are not seen in pre-training, as the support/query sets.

\vspace{0.1em}

\noindent \textbf{2) Comparison and Analytics.}
The results are summarized in Table~\ref{tab:other_tasks}. We utilize the officially released model weights, keeping the experiments as fair and aligned as possible.

We first present the discussion on the perspective of model backbones. GRU models usually employ higher representation dimensions due to the smaller GPU memory consumption, which makes a more informative representation available, leading to stronger performance in retrieval task than GCNs and Transformers. GCNs model the skeleton more compactly with structured prior knowledge, resulting in a better capacity to handle over-fitting, accounting for the stronger few-shot performance for unseen categories. However, the other side of the coin is the sensitivity to occlusion, where GCN can suffer from serious representation degradation because of the distribution shift caused by occlusion. Besides, it is found the GRU-based often performs better than the GCN model in terms of action detection, which primarily focuses on the spatial structure while ignoring the fine-grained temporal dependency. In contrast, GRUs model the skeletons in a temporal-based auto-regressive manner and possess better scalability for the long-sequences and short-clips. Besides, the Transformer-based model~\cite{mao2023masked} also demonstrates promising results when employing the masked skeleton modeling pretext task, and can be expected with a better fine-tuning performance due to the powerful attention mechanism.   

On the taxonomy of pretext tasks, due to the lack of promising open-sourced context-based methods, we mainly focus on contrastive and generative learning methods. First, it is found that contrastive learning methods usually perform better than reconstruction-based methods, \eg, MAMP~\cite{mao2023masked} and CRRL~\cite{qi2023contrast}, in action retrieval and few-shot learning. These tasks mainly rely on the direct measure in the representation space, \ie, cosine similarity, indicating the dominant position of contrastive learning for achieving highly compact representations. 
Remarkably, the masked modeling method, \eg, MAMP~\cite{mao2023masked}, achieves significant results in action detection task. We believe that masked modeling with Transformer architecture is very promising for handling the action detection task. By splitting the long sequences into short clips as patches for masked prediction, the learned representations can well capture fine-grained action cues and achieve notable results. However, it is not observed with as good performance  under occluded recognition as expected, although it explicitly involves masked data modeling, which can be explored in future work.

\subsubsection{\revise{Cross-Dataset Analytics for Different Downstream Tasks}}
We study the effect of the dependency relationship between different pre-training datasets on multiple downstream tasks with the proposed method. Specifically, we pre-train the model on NTU 60, NTU 120, PKUMMD, and their hybrid dataset, respectively, which is then evaluated on various downstream tasks under the transfer learning setting. To achieve a better understanding, we first introduce the relationship between different datasets in terms of the data composition. Concretely, NTU 120 extends NTU 60 on the action categories from 60 to 120 without introducing new samples of the original categories. PKUMMD is another skeleton dataset with similar collection settings, sharing a significant overlap of category distribution with NTU 60 and possessing fewer samples, which can be approximated as a smaller version of NTU 60. The quantitative results are presented in Table~\ref{tab:dataset_dependency}. 

\noindent$\bullet$ First, as expected, reducing the scale of pre-training data generally results in performance degradation for downstream tasks, especially for the evaluation of the large-scale NTU 120 dataset. This is because the new classes in testing are unseen during pre-training, resulting in limited generalization capacity.

\noindent$\bullet$ Second, when the data distribution in testing is highly consistent with pre-training, introducing additional pre-training data of new categories (\eg, from NTU 60 to NTU 120) is not as effective as increasing the sample number within the original categories (\eg, from NTU 120 to NTU 120+PKUMMD, from PKUMMD to NTU 60). Moreover, it is even observed with slight performance degradation, \eg, retrieval, spatial occluded recognition, and detection tasks (75.4$\rightarrow$75.3, 88.0$\rightarrow$87.7, 75.5$\rightarrow$75.3). This can be interpreted as the new action category learning occupies part of the model capacity, impairing the representations of original categories.

\noindent$\bullet$ In contrast, for the tasks requiring to explicitly generalize to new data distributions, \eg, temporal occluded recognition with distribution shift by impairing data structure (Note only spatial occlusion is seen in the pre-training) and few shot tasks with new categories, introducing new categories can benefit the performance significantly. This indicates more essential knowledge of human motion is enabled for model to boost generalization.

To sum up, we analyze the effect of pre-training on different datasets to downstream tasks from two factors, category distribution and the sample number within each class, providing valuable insights for the skeleton SSL dataset choice and construction. Overall, a better performance can be expected with a larger pre-training dataset. However, it is also necessary to consider the actual downstream scenarios and the adopted pre-training methods to adjust data component.

\begin{table}[t]
  \centering
  \caption{{Ablation study on the novel positive pair and semantic guidance.}}
  \begin{tabular}{cc|c|c}
   \toprule
   \multicolumn{2}{c|}{Novel Positive Pair} & \multirow{2.5}{*}{Semantic Guidance} & \multirow{2.5}{*}{xview}\\
   \cmidrule{1-2}
   $s_{mask}$  &$s_{predict}$ &  &\\
   \midrule
    &  & & 88.2\\
   \checkmark & & & 89.3\\
   \checkmark & \checkmark & &90.0\\
   \checkmark & \checkmark & \checkmark & \textbf{90.5}\\
    \bottomrule
\end{tabular}
  \label{tab:novel_pos}
  \vspace{-5pt}
\end{table}

\begin{table}[t]
			\caption{Ablation study on the clip-level contrastive learning and the post-distillation.}
			\centering
			\vspace{-5pt}
   \setlength{\tabcolsep}{5pt}
			  \begin{tabular}{l|cc|c}
   \toprule
    \multirow{2}{*}{Method} & \multicolumn{2}{c|}{Recognition} &Detection\\
    & xsub & xview & Part I\\
    \midrule
    \textit{w/o} clip contrastive learning & 83.9 &90.4 & 73.3\\
    \textit{w} clip contrastive learning &\textbf{84.1} & \textbf{90.8} & \textbf{75.5}\\
    \midrule
    \textit{w/o} post-distillation & 84.1 & 90.8 & {75.5} \\
    \textit{w} post-distillation & \textbf{84.8} & \textbf{91.0} & 75.5\\
    \bottomrule
\end{tabular}
  \label{tab:prompt}
	\end{table}	
    

\subsection{{Analysis of PCM$^{3}$++}}
\label{sec:ablation}

\noindent \textbf{1) Evaluating the PCM$^{3}$++.}
As shown in the previous results in Table~\ref{tab:main_comp} and Table~\ref{tab:other_tasks}, the proposed method achieves remarkable results on different downstream tasks compared with the state-of-the-art methods, and demonstrates significant improvement over our baseline method~\cite{zhang2023prompted}. As our new improvement, we integrate an effective clip-level contrastive learning scheme and present a novel post-distillation training strategy. These methodological advancement is studied as follows. 

\noindent \textbf{2) Analysis of clip contrastive learning.} 
We add the temporal clips as the asymmetric positive sample of the anchor sequence, which further improves the representation quality as shown in Table~\ref{tab:prompt}. Remarkably, the clip-level contrastive learning significantly boost the action detection performance by promoting short-term modeling capacity. Our designs boost versatile representation learning of different granularity and the effect on more downstream tasks can be found in Supp.

\noindent \textbf{3) Effect of the post-distillation design.}
As shown in Table~\ref{tab:prompt}, the post-distillation training strategy can bring further improvement slightly for the recognition task. It alleviates the false negative problem by removing the one-hot pseudo label in the contrastive learning, achieving a more compact representation space. 

\begin{table}[t]
  \centering
    \color{black}
  \caption{FLOPs, params, and training time (all tested on a single NVIDIA A40 GPU) results.}
  \vspace{-5pt}
  \begin{tabular}{l|ccc|c}
   \toprule
   \multirow{2}{*}{Model} & \multirow{2}{*}{Params} &\multirow{2}{*}{FLOPs} & {Training} &\multirow{2}{*}{Acc.} \\
    & & & Time & \\
    \midrule
    GL-Trans.~\cite{kim2022global}	&214M	&59.4G &53.62h &83.8\%\\
    ISC~\cite{thoker2021skeleton}	&106M & 13.7G & 45.49h & 85.2\% \\
    CMD~\cite{mao2022cmd} &99M &17.3G& 44.13h &86.9\% \\
    \midrule
    PCM$^{3}$~\cite{zhang2023prompted} & 103M  &15.0G & 45.31h &{90.4\%} \\
    PCM$^{3}$++ & 103M  &17.9G &  47.95h &\textbf{91.0\%} \\
    \bottomrule
\end{tabular}
  \vspace{-5pt}\label{tab:abl_flops}
\end{table}



\noindent\revise{\textbf{4) Novel positive pair and semantic guidance.} We analyze the effect of synergetic design between masked prediction and contrastive learning in Table~\ref{tab:novel_pos}. As we can see, the masked views and predicted views as positives bring notable improvement, indicating the potential benefits of masked prediction training to contrastive learning. On the other hand, we propagate the gradients of contrastive branch to masked prediction as high-level semantic guidance rather than stop the gradient. This boosts masked prediction training and yields more informative positive samples, leading to further performance improvement.}

\noindent \textbf{5) Complexity analysis.}
We give an analysis of space and computational complexities of our method for pre-training in Table~\ref{tab:abl_flops}. Compared with other GRU-based and Transformer-based methods, our method achieves a significant performance improvement with an acceptable cost of the complexity and training time. For the space complexity, the main extra cost is the reconstruction decoder (4M of space). For the computational complexity, the encoding process of different positive samples contributes most of the overhead.

\section{Conclusions and Future Directions}
\label{sec:conclusion}
This paper presents a comprehensive survey on skeleton-based action representation SSL, where different literature is organized following
the taxonomy of context-based, generative learning, and contrastive learning approaches. After reviewing existing works, we technically propose a novel and effective framework, for versatile skeleton-based action representation learning which is less explored before as a challenging topic. Finally, a detailed benchmark and insightful discussions are provided, including datasets, model backbones and pretext tasks. Meanwhile, we demonstrate the superior generalization performance of our method under different downstream tasks.

For the future research, the following pending issues deserve more attention:
\begin{itemize}[leftmargin=1em]

\item \textbf{Long-Term Motion Understanding.} Existing methods utilize the trimmed video clip, only containing one motion, as training data. This limits the long-term temporal reasoning capacity containing multiple actions, \eg, \textit{Long-Term Action Anticipation} task.

\item \textbf{Multi-Modal Learning.} \revise{The exploration of multi-modal pre-training including human skeleton data is still insufficient, \eg, texts, and RGB images. For example, as a useful complement to skeleton, RGB data can provide additional background knowledge that can boost action representation learning. Some recent works~\cite{petrovich2023tmr,yu2024mopatch} explore the human motion-text representation learning. However, the insufficiency of high-quality motion-text datasets still limits the model generalization capacity.}

\item \textbf{Towards Skeleton in the Wild.} Existing methods are evaluated in simplified and controlled environments and can suffer from serious noise, \eg, occlusion and view variation when deployed on the outdoor scenario. On this point, SSL is promising to boost skeleton representation robustness in the wild.

\item \textbf{Versatile Representation Learning.} As discussed in previous sections, most works only focus on the recognition task, leaving the generation capacity across different tasks of SSL models under-explored. Meanwhile, new frameworks can also be explored, \eg, Diffusion model~\cite{ho2020denoising}, which is promising to handle both generative and discriminative downstream tasks.

\end{itemize}







\section*{Declarations}
\begin{itemize}
\item Competing interests. {This work was supported in part by the Beijing Major Science and Technology Project under Contract No. Z251100008425023, in part by the National Natural Science Foundation of China under Grant 62471009, and in part by the Key Laboratory of Science, Technology and Standard in Press Industry (Key Laboratory of Intelligent Press Media Technology).}
\item Data availability. {The adopted NTU-RGB+D~\cite{shahroudy2016ntu,liu2019ntu} and PKU-MMD~\cite{liu2020pku}
dataset are popular public
benchmarks in skeleton-based action understanding. The
code for data processing can be found in https://github.com/JHang2020/PCM3.}

\end{itemize}

\bibliographystyle{plain}
\footnotesize
\bibliography{ref}

\end{document}